\documentclass{article}

\usepackage[utf8]{inputenc}
\usepackage{soul}
\usepackage{xspace}
\usepackage{amsmath,amsthm,amsfonts, amssymb}
\usepackage{subcaption}
\usepackage{tikz}
\usepackage{tikz-qtree}
\usetikzlibrary{patterns}
\usetikzlibrary{shapes.geometric}
\usetikzlibrary{positioning}
\usetikzlibrary{arrows,automata,calc,backgrounds,fit}

\newtheorem{definition}{Definition}
\newtheorem{proposition}{Proposition}
\newtheorem{lemma}{Lemma}
\newtheorem{corollary}{Corollary}
\newtheorem{example}{Example}
\newtheorem{theorem}{Theorem}

\tikzset{
    >=stealth',
    args/.style={circle,draw=black,minimum size=0.85cm}    
}

\newcommand{\AF}{\ensuremath{\mathsf{AF}}\xspace}						
\newcommand{\arguments}{\ensuremath{\mathsf{A}}\xspace}				
\newcommand{\attacks}{\ensuremath{\mathsf{R}}\xspace}				
\newcommand{\AFcomplete}{\ensuremath{\AF=(\arguments,\attacks)}\xspace}		

\newcommand{\allAFs}{\ensuremath{\mathfrak{AF}}}
\newcommand{\allArgs}{\ensuremath{\mathfrak{A}}}

\newcommand{\adm}{\textsf{adm}}
\newcommand{\co}{\textsf{co}}
\newcommand{\gr}{\textsf{gr}}
\renewcommand{\st}{\textsf{st}}
\newcommand{\pr}{\textsf{pr}}
\newcommand{\sst}{\textsf{sst}}
\newcommand{\id}{\textsf{id}}
\newcommand{\sa}{\textsf{sa}}

\newcommand{\elc}{\textsf{IS}}
\newcommand{\uelc}{\ensuremath{\textsf{IS}^{\not\leftarrow}}}
\newcommand{\ucelc}{\ensuremath{\textsf{IS}^{\not\leftrightarrow}}}
\newcommand{\celc}{\ensuremath{\textsf{IS}^{\leftrightarrow}}}
\newcommand{\conflicts}{\textsf{conflicts}}

\newcommand{\SCC}{\textsf{SCC}}
\newcommand{\SCCs}{\textsf{SCCs}}

\title{Revisiting initial sets in abstract argumentation}
\author{Matthias Thimm}
\date{Artificial Intelligence Group, University of Hagen, Germany\\[1ex]\texttt{matthias.thimm@fernuni-hagen.de}}

\begin{document}
\maketitle

\begin{abstract}
We revisit the notion of \emph{initial sets} by Xu and Cayrol \cite{Xu:2016}, i.\,e., non-empty minimal admissible sets in abstract argumentation frameworks. Initial sets are a simple concept for analysing conflicts in an abstract argumentation framework and to explain why certain arguments can be accepted. We contribute with new insights on the structure of initial sets and devise a simple non-deterministic construction principle for any admissible set, based on iterative selection of initial sets of the original framework and its induced reducts. In particular, we characterise many existing admissibility-based semantics via this construction principle, thus providing a constructive explanation on the structure of extensions. We also investigate certain problems related to initial sets with respect to their computational complexity.
\end{abstract}


\section{Introduction}
\emph{Formal argumentation} \cite{Atkinson:2017,Baroni:2018} encompasses approaches for non-monotonic reasoning that focus on the role of arguments and their interactions. The most well-known approach is that of \emph{abstract argumentation frameworks} \cite{Dung:1995} that model arguments as vertices in a directed graph, where a directed edge from an argument $a$ to an argument $b$ denotes an \emph{attack} from $a$ to $b$. Although conceptually simple, abstract argumentation frameworks can be used in a variety of argumentative scenarios such as persuasion dialogues \cite{DBLP:conf/comma/ChalaguineH20}, explanations in recommendation systems \cite{DBLP:conf/kr/0001CBT20}, mathematical modelling \cite{DBLP:conf/comma/BoudjaniGK18}, or as a target formalism of \emph{structured argumentation formalisms} \cite{Modgil:2014,DBLP:journals/argcom/Toni14}. Formal semantics are given to abstract argumentation frameworks by \emph{extensions}, i.\,e., sets of arguments that can jointly be accepted and represent a coherent standpoint on the conflicts between the arguments. Different variants of such semantics have been proposed \cite{Baroni:2018a}, but there are also other (non set-based) approaches such as ranking-based semantics \cite{Amgoud:2013} and probabilistic approaches \cite{Hunter:2017}. 

A particular advantage of approaches to formal argumentation is the capability to \emph{explain} the reasoning behind certain conclusions using human-accessible concepts such as arguments and counterarguments.
Works such as \cite{Amgoud:2009,DBLP:conf/kr/0001CBT20,DBLP:journals/synthese/SeseljaS13,Liao:2020,Saribatur:2020,DBLP:conf/kr/NiskanenJ20,Ulbricht:2021} have already explored certain aspects of the explanatory power of approaches to formal argumentation. Amgoud and Prakken \cite{Amgoud:2009} and Rago et.\ al \cite{DBLP:conf/kr/0001CBT20} develop argumentation formalisms for decision-making that augment recommendations with arguments. In \cite{DBLP:journals/synthese/SeseljaS13} an extension to abstract argumentation frameworks is developed that explicitly includes a relationship for an ``explanation'', while Liao and van der Torre \cite{Liao:2020} define ``explanation semantics'' for ordinary abstract argumentation frameworks. Saribatur, Wallner, and Woltran \cite{Saribatur:2020} as well as Niskanen and J\"arvisalo \cite{DBLP:conf/kr/NiskanenJ20} address computational problems and develop a notion for explaining non-acceptability of arguments to, e.\,g., verify results of an argumentation solver. Finally, \cite{Ulbricht:2021} presents \emph{strong explanations} as a mechanism to explain acceptability of (sets of) arguments.

In this paper, we revisit one of the fundamental concepts underlying approaches to formal argumentation (and abstract argumentation in particular) for the purpose of explaining, namely \emph{admissibility}. Informally speaking, a set of arguments is \emph{admissible} if each of its members is defended against any attack from the outside (we will provide formal details in Section~\ref{sec:aa}). Many popular semantics for abstract argumentation rely on the notion of admissibility. In particular, a \emph{preferred extension} is a maximal (wrt.\ set inclusion) admissible set and preferred semantics satisfies many desirable properties \cite{Baroni:2018a}. However, since a preferred extension is a maximal admissible set, it can hardly be used for explaining why a certain argument is acceptable: such an extension may contain many irrelevant arguments and its size alone distracts from the particular reasons why a certain member is acceptable. 
Our aim is to investigate why certain arguments are contained in, e.\,g., a preferred extension and how we can decompose such large extensions into smaller sets that allow us to justify the reasoning process behind such complex semantics.
As a tool for our investigation, we consider \emph{initial sets}, i.\,e., non-empty admissible sets that are minimal wrt.\ set inclusion.
Initial sets have been introduced in \cite{Xu:2016} and further analysed in \cite{Xu:2018,Xu:2018a}. We contribute to this analysis with new insights on the structure of initial sets and, in particular, to the use of initial sets for the task of explanation.
In fact, initial sets can exactly be used for the purpose mentioned before \cite{Xu:2016}: they allow us to decompose large extensions into smaller fragments, each of them representing a single resolved issue in the argumentation framework and thus showcases the reasoning behind certain semantics. This has been done already in some form in \cite{Xu:2016,Xu:2018,Xu:2018a} but we present a new, and arguably more elegant, formalisation of that idea that allows us to derive new results as well. Using the notion of a \emph{reduct} \cite{Baumann:2020}, we can concisely represent any admissible set as a sequence of initial sets of the original framework and derived reducts. Moreover, we characterise many admissibility-based semantics through a step-wise construction process using certain selections of initial sets (this has been hinted at using the original formalisation for complete and preferred semantics in \cite{Xu:2018}). We round up our analysis with a characterisation of the computational complexity of certain tasks related to initial sets, which is also missing so far from the literature.

In summary, the contributions of this paper are as follows:
\begin{enumerate}
	\item We revisit initial sets and investigate further properties (Section~\ref{sec:cores})
	\item We provide a characterisation result of admissible sets and many admissibility-based semantics (Section~\ref{sec:cores2})
	\item We analyse certain computational problems wrt.\ their complexity (Section~\ref{sec:complexity})
\end{enumerate}
Section~\ref{sec:aa} introduces preliminaries on abstract argumentation and Section~\ref{sec:summary} concludes this paper. 

Complete proofs can be found in the appendix. A short paper presenting the main ideas of this work has been published before in \cite{Thimm:2021c}.

\section{Abstract Argumentation}\label{sec:aa}
Let $\allArgs$ denote a universal set of arguments. An \emph{abstract argumentation framework} $\AF$ is a tuple $\AF=(\arguments,\attacks)$ where $\arguments\subseteq \allArgs$ is a finite set of arguments and \attacks is a relation $\attacks\subseteq \arguments\times\arguments$ \cite{Dung:1995}. Let $\allAFs$ denote the set of all abstract argumentation frameworks.
 For two arguments $a,b\in\arguments$ the relation $a \attacks b$ means that argument $a$ attacks argument $b$. 
 For $\AFcomplete$ and $\AF'=(\arguments',\attacks')$ we write $\AF'\sqsubseteq\AF$ iff $\arguments'\subseteq \arguments$ and $\attacks'=\attacks\cap(\arguments'\times\arguments')$. For a set $X\subseteq \arguments$, we denote by $\AF|_X=(X,\attacks\cap(X\times X))$ the projection of $\AF$ on $X$.
For a set $S\subseteq \arguments$ we define
\begin{align*}
    S^+ & = \{a\in \arguments\mid \exists b\in S: b\attacks a\} \\
    S^- & = \{a\in \arguments\mid \exists b\in S: a\attacks b\}
\end{align*}
If $S$ is a singleton set, we omit brackets for readability, i.\,e., we write $a^{-}$ ($a^{+}$) instead of $\{a\}^{-}$ ($\{a\}^{+}$). For two sets $S$ and $S'$ we write $S\attacks S'$ iff $S^+\cap S'\neq \emptyset$.
We say that a set $S\subseteq\arguments$ is \emph{conflict-free} if for all $a,b\in S$ it is not the case that $a\attacks b$.
A set $S$ \emph{defends} an argument $b\in\arguments$ if for all $a$ with $a\attacks b$ there is $c\in S$ with $c\attacks a$.
A conflict-free set $S$ is called \emph{admissible} if $S$ defends all $a\in S$. Let $\adm(\AF)$ denote the set of admissible sets of $\AF$.

Different semantics can be phrased by imposing constraints on admissible sets \cite{Baroni:2018a}. 
In particular, an admissible set $E$
\begin{itemize}
   	 \item is a \emph{complete} (\co) extension iff for all $a\in \arguments$, if $E$ defends $a$ then $a\in E$,
	\item is a \emph{grounded} (\gr) extension iff $E$ is complete and minimal,
	\item is a \emph{stable} (\st) extension iff $E\cup E^+ =\arguments$,
	\item is a \emph{preferred} (\pr) extension iff $E$ is maximal.
	\item is a \emph{semi-stable} (\sst) extension iff $E\cup E^+$ is maximal,
	\item is an \emph{ideal} (\id) extension iff $E$ is the maximal admissible set with $E\subseteq E'$ for each preferred extension $E'$.
	\item is a \emph{strongly admissible} (\sa) extension iff $E=\emptyset$ or each $a\in E$ is defended by some strongly admissible $E'\subseteq E\setminus\{a\}$.
\end{itemize}
All statements on minimality/maximality are meant to be with respect to set inclusion. For $\sigma\in\{\co,\gr,\st,\pr,\sst,\id,\sa\}$ let $\sigma(\AF)$ denote the set of $\sigma$-extensions of $\AF$. We say a semantics $\sigma$ is \emph{admissibility-based} if $\sigma(\AF)\subseteq\adm(\AF)$ for all $\AF$. Note that all semantics above are admissibility-based but there are also others such as CF2 semantics \cite{Baroni:2005a} and weak admissibility-based semantics \cite{Baumann:2020}.

\section{Revisiting initial sets}\label{sec:cores}
Admissibility captures the basic intuition for an explanation \emph{why} a certain argument can be regarded as acceptable. More concretely, if $S$ is an admissible set then $a\in S$ is accepted \emph{because} all arguments in $S$ are accepted, every attacker of $a$ is attacked back by some argument in $S$. However, admissibility alone is not sufficient to model explainability as it does not take \emph{relevance} into account. 
\begin{example}\label{ex:ex1}
	Consider the argumentation framework $\AF_0$ depicted in Figure~\ref{fig:exex1}. There are eight admissible sets containing the argument $e$:
	\begin{align*}
		 S_1 & = \{b,e,f,h,i\} & S_2 & = \{b,e,f,i\}  & S_3 & = \{b,e,h,i\} \\
		 S_4 & = \{e,f,h,i\} & S_5 & = \{b,e,i\} & S_6 & = \{f,e,i\}\\
		 S_7 & = \{h,e,i\} &  S_8 & = \{e,i\}
	\end{align*}
	$S_1$ is also a preferred extension. However, it is also clear that arguments $b$, $f$, and $h$ are not integral for defending $e$ and the set $S_8$ presents a concise description of what is needed in order to deem $e$ as acceptable (wrt.\ admissibility), namely only $e$ and $i$.
\begin{figure}[t]
\begin{center}
\begin{tikzpicture}[scale=0.8, every node/.style={scale=0.8},node distance=0.7cm]

	\node[args](a){$a$};
	\node[args, right=of a](b){$b$};
	\node[args, right=of b](c){$c$};
	\node[args, right=of c](d){$d$};
	\node[args, right=of d](e){$e$};
	\node[args, below=of a](f){$f$};
	\node[args, right=of f](g){$g$};
	\node[args, right=of g](h){$h$};
	\node[args, right=of h](i){$i$};
	\node[args, right=of i](j){$j$};
	
	\path(a) edge [->,loop,in=320,out=30,min distance=10mm] (a);
	\path(f) edge [->,bend left] (g);
	\path(g) edge [->,bend left] (f);
	\path(a) edge [->,bend left] (f);
	\path(f) edge [->,bend left] (a);
	\path(g) edge [->] (b);
	\path(c) edge [->] (b);
	\path(h) edge [->] (g);
	\path(d) edge [->] (c);
	\path(i) edge [->,bend left] (g);
	\path(i) edge [->] (j);
	\path(j) edge [->] (e);
	\path(e) edge [->] (d);
	\path(d) edge [->] (i);
	\path(i) edge [->] (c);

\end{tikzpicture}
\end{center}
\caption{$\AF_0$ from Example~\ref{ex:ex1}.}
\label{fig:exex1}
\end{figure}
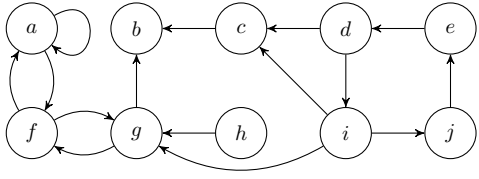
\end{example}
In the following, we take relevance into account by considering \emph{minimal} (wrt.\ set inclusion) admissible sets.
Of course, a notion of minimal admissible set without further constraints is not a useful concept as the empty set is always admissible and constitutes the unique minimal admissible set. 
Non-empty minimal admissible sets have been coined \emph{initial sets} by Xu and Cayrol in \cite{Xu:2016}.
\begin{definition}[\cite{Xu:2016}]
	For $\AFcomplete$, a set $S\subseteq\arguments$ with $S\neq\emptyset$ is called an \emph{initial set} if $S$ is admissible and there is no admissible $S'\subsetneq S$ with $S'\neq \emptyset$. Let $\elc(\AF)$ denote the set of initial sets of $\AF$.
\end{definition}
\begin{example}\label{ex:ex3}
	We continue Example~\ref{ex:ex1}. There are four initial sets of $\AF_0$: $\{f\}$, $\{h\}$, $\{d,j\}$, and $\{e,i\}$.
\end{example}
As the previous example shows, an initial set is not supposed to provide a ``solution'' to the whole argumentation represented in an abstract argumentation framework, but ``solves'' a single atomic conflict (or in the case of $\{h\}$ points to an obvious deterministic inference step). In fact, we can identify three different types of initial sets.
\begin{definition}
	For $\AFcomplete$ and $S\in\elc(\AF)$, we say that
	\begin{enumerate}
		\item $S$ is \emph{unattacked} iff $S^{-}=\emptyset$,
		\item $S$ is \emph{unchallenged} iff $S^{-}\neq\emptyset$ and there is no $S'\in \elc(\AF)$ with $S'\attacks S$,
		\item $S$ is \emph{challenged} iff there is $S'\in \elc(\AF)$ with $S'\attacks S$.
	\end{enumerate}
\end{definition}
Note that only unattacked initial sets have been considered explicitly in \cite{Xu:2018}; in particular, note that every unattacked initial set $S$ is a singleton $S=\{a\}$. Observe that the notions of unattacked, unchallenged, and challenged initial sets are mutually exclusive and exhaustive. Let $\uelc(\AF)$, $\ucelc(\AF)$, and $\celc(\AF)$ denote the set of unattacked, unchallenged, and challenged initial sets, respectively. So we have $\elc(\AF)=\uelc(\AF)\cup \ucelc(\AF)\cup \celc(\AF)$.
Moreover, for $S\in \elc(\AF)$ let
\begin{align*}
	\conflicts(S,\AF) & = \{S'\in \elc(\AF) \mid S'\attacks S\}
\end{align*}
denote the set of conflicting initial sets of $S$, which is always empty in the case of unattacked and unchallenged initial sets. Note that $S\attacks S'$ implies $S'\attacks S$ for any $S,S'\in \elc(\AF)$ as $S'$ is admissible and therefore defends itself. 
\begin{example}
	We continue Example~\ref{ex:ex3}. Here we have
	\begin{align*}
		\uelc(\AF) & = \{\{h\}\}\\
		\ucelc(\AF) & = \{\{f\}\}\\
		\celc(\AF) & = \{\{d,j\},\{e,i\}\}
	\end{align*}
	and $\{d,j\}$ and $\{e,i\}$ are in conflict with each other, i.\,e., $\conflicts(\{d,j\},\AF)=\{\{e,i\}\}$ and $\conflicts(\{e,i\},\AF)=\{\{d,j\}\}$.
\end{example}
Before we continue with characterising arbitrary admissible sets using initial sets in Section~\ref{sec:cores2}, we first contribute some new results on the structure of initial sets, therefore extending the analysis from \cite{Xu:2016,Xu:2018,Xu:2018a}.


Initial sets have an interesting property with respect to strongly connected components as follows. Recall that we can decompose an abstract argumentation framework $\AF$ into its strongly connected components. More precisely, an abstract argumentation framework $\AF'=(\arguments',\attacks')$ is a strongly connected component (\textsf{SCC}) of $\AF$, if $\AF'\sqsubseteq\AF$ s.t.\ there is a directed path between any pair $a,b\in\arguments'$ in $\AF'$ and there is no larger $\AF''$ with that property. Let $\SCC(\AF)$ be the set of \SCCs\ of \AF.
\begin{example}
	Consider again the argumentation framework $\AF_0$ from Figure~\ref{fig:exex1}. $\AF_0$ decomposes as follows into \SCCs:
	$\SCC(\AF_0)  = \{\AF_0|_{\{d,e,i,j\}},\AF_0|_{\{c\}},$\\ $\AF_0|_{\{h\}},\AF_0|_{\{g,f,a\}},\AF_0|_{\{b\}}\}$
\end{example}
The following result shows that initial sets are always completely contained in a single \SCC.
\begin{proposition}\label{prop:scc1}
	If $S$ is an initial set of $\AF$ then there is $\AF'=(\arguments',\attacks')\in\SCC(\AF)$ s.t.\ $S\subseteq \arguments'$.
\end{proposition}
If $S$ is an initial set let $\SCC(S)$ denote its \SCC\ as in the above proposition. Initial sets can actually be characterised by their \SCC\ as follows.
\begin{proposition}\label{prop:scc2}
	$S$ is an initial set of $\AF$ if and only if $S$ is an initial set of $\SCC(S)=(\arguments',\attacks')$ and $S^{-}\subseteq \arguments'$.
\end{proposition}
In other words, a set $S$ is an initial set iff it is an initial set of a single \SCC\ and it is not attacked by arguments outside of the \SCC.

We close this investigation on the relationship of initial sets with \SCCs\ by making some straightforward observations regarding the types of initial sets.
\begin{proposition}
	Let $S\in\elc(\AF)$ and $\SCC(S)=(\arguments',\attacks')$.
	\begin{enumerate}
		\item If $S$ is unattacked then $|\arguments'|=1$.
		\item If $S$ is challenged or unchallenged then $|\arguments'|>1$.
		\item If $S$ is challenged and $S'\in\conflicts(S,\AF)$ then $\SCC(S)=\SCC(S')$.
	\end{enumerate}
\end{proposition}
In particular, the final observation in the previous proposition states that conflicts between initial sets are always within a single \SCC.
\section{Characterising Admissibility-based Semantics through Initial Sets}\label{sec:cores2}
In \cite{Xu:2016,Xu:2018} it has been shown that any admissible set (and in particular every complete and preferred extension) can be constructed by 1.) selecting a set of non-conflicting initial sets, 2.) adding further defended arguments, and 3.) iterating this procedure taking so-called ``J-acceptable'' sets into account. In particular, the described mechanism involves iterative application of the characteristic function \cite{Dung:1995}, computation of the grounded extension, and said notion of J-acceptability to provide those characterisations (and some further concepts). In this section, we provide a (arguably) more elegant formalisation of these ideas that allows us to derive characterisations of further semantics as well as some impossibility results. Results that are (partly) due to works \cite{Xu:2016,Xu:2018,Xu:2018a} are annotated as such, all remaining results are new.

Our characterisations rely on the notion of the \emph{reduct} \cite{Baumann:2020}.
\begin{definition}
	For $\AFcomplete$ and $S\subseteq \arguments$, the $S$-reduct $\AF^{S}$ of $\AF$ is defined via $\AF^{S}=\AF|_{\arguments\setminus(S\cup S^{+})}$.
\end{definition}
As a single initial set $S$ solves an atomic conflict in an abstract argumentation framework $\AF$, ``committing'' to it by moving to $\AF^{S}$ may reveal further conflicts and thus new initial sets. We can make the following observations on this aspect.
\begin{proposition}
	Let  $\AFcomplete$ be an abstract argumentation framework and $S,S'\in \elc(\AF)$ with $S\neq S'$.
	\begin{enumerate}
		\item If $S'\in \uelc(\AF)$ then $S'\in \uelc(\AF^{S})$
		\item If $S'\in \conflicts(S,\AF)$ then $S'\notin \elc(\AF^{S})$
		\item If $S'\notin\conflicts(S,\AF)$ then $S'\cap \bigcup\elc(\AF^{S})\neq \emptyset$
	\end{enumerate}
\end{proposition}
The above observations give an impression on how initial sets behave under reducts. So unattacked initial sets are always retained (item 1), conflicting initial sets are always removed (item 2), and non-conflicting initial sets are ``essentially'' retained (item~3). More precisely, while it may not be guaranteed that non-conflicting initial sets are still initial sets in the reduct, their arguments are still potentially acceptable, as the following example shows.
\begin{example}\label{ex:cores:red}
	Consider the argumentation framework $\AF_{1}$ depicted in Figure~\ref{fig:cores:red}. We have
	\begin{align*}
		\elc(\AF_{1})  &= \{\{a,c\},\{b,d\},\{e\}\}
	\end{align*}
	and $\{b,d\}$ and $\{e\}$ are conflicting (there are no further conflicts). Now we have
	\begin{align*}
		\elc(\AF_{1}^{\{e\}} ) & = \{\{c\}\}
	\end{align*}
	So the initial set $\{a,c\}$ of $\AF_{1}$ is not retained in $\AF_{1}^{\{e\}}$, despite $\{a,c\}$ and $\{e\}$ not being in conflict. However, we have that $\{c\}\subseteq\{a,c\}$ is an initial set of $\AF_{1}^{\{e\}}$. Furthermore, $\{a\}$ (the ``remaining'' argument of the initial set $\{a,c\}$) is actually the unique initial set of $(\AF_{1}^{\{e\}})^{\{c\}}$.
\begin{figure}[t]
\begin{center}
\begin{tikzpicture}[scale=0.8, every node/.style={scale=0.8},node distance=0.7cm]

	\node[args](a){$a$};
	\node[args, right=of a](b){$b$};
	\node[args, below=of b](c){$c$};
	\node[args, below=of a](d){$d$};
	\node[args, right=of b](e){$e$};
	
	\path(a) edge [->] (b);
	\path(b) edge [->] (c);
	\path(c) edge [->] (d);
	\path(d) edge [->] (a);
	\path(b) edge [->, bend left] (e);
	\path(e) edge [->, bend left] (b);
\end{tikzpicture}
\end{center}
\caption{$\AF_1$ from Example~\ref{ex:cores:red}.}
\label{fig:cores:red}
\end{figure}
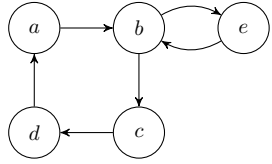
\end{example}
The following results show that by an iterative selection of initial sets on the corresponding reducts, we can re-construct every admissible set.
These observations have been hinted at in \cite{Xu:2016,Xu:2018,Xu:2018a}, but no formal proof had been provided. We make up for that now.
\begin{theorem}\label{th:elc1}
	Let  $\AFcomplete$ be an abstract argumentation framework and $S\subseteq \arguments$. $S$ is admissible if and only if either
	\begin{itemize}
		\item $S=\emptyset$ or
		\item $S=S_{1}\cup S_{2}$, $S_{1}\in\elc(\AF)$ and $S_{2}$ is admissible in $\AF^{S_{1}}$.
	\end{itemize}
	\begin{proof}
		Let $S$ be an admissible set and $S\neq \emptyset$. By definition of initial sets, there is $S_{1}\in \elc(\AF)$ with $S_{1}\subseteq S$. It remains to show that $S_{2}=S\setminus S_{1}$ is admissible in $\AF^{S_{1}}$. Let $a\in S_{2}$ and let $b_{1},\ldots, b_{n}\in\arguments$ be the attackers of $a$ in $\AF$. Since $S$ is admissible, there are arguments $c_{1},\ldots,c_{n}\in S$ so that $c_{i}$ attacks $b_{i}$, $i=1,\ldots,n$ (possibly some of the $c_{i}$ are identical). Without loss of generality, assume $c_{1},\ldots, c_{k}\in S_{1}$ for some $k\leq n$. Then $b_{1},\ldots, b_{k}$ are not present in $\AF^{S_{1}}$, thus $a$ must only be defended against $b_{k+1},\ldots, b_{n}$  in $\AF^{S_{1}}$. However, since $S_{2}=S\setminus S_{1}$ we have that $c_{k+1},\ldots,c_{n}\in S_{2}$, showing that $a$ is defended by $S_{2}$ in $\AF^{S_{1}}$ and, thus, $S_{2}$ is admissible in $\AF^{S_{1}}$. 
		
		 For the other direction\footnote{Note that this direction can also be shown by using the fact that admissible ``semantics'' is fully decomposable \cite{Baroni:2014b}, but we explicitly prove it for matters of simplicity.}, if $S=\emptyset$ then $S$ is also admissible. Assume $S=S_{1}\cup S_{2}$, $S_{1}\in\elc(\AF)$ and $S_{2}$ is an admissible set of $\AF^{S_{1}}$. We have to show that $S$ is admissible. Let $a\in S$ and let $b_{1},\ldots, b_{n}$ be the attackers of $a$ in $\AF$. If $a\in S_{1}$ then there are $c_{1},\ldots,c_{n}\in S_{1}\subseteq S$ such that $c_{i}$ attacks $b_{i}$ since $S_{1}$ is admissible. If $a\in S_{2}$, assume for the sake of contradiction that there is an attacker $b$ of $a$ such that there is no $b\in S$ that attacks $c$ in $\AF$. It follows that $b$ is also in $\AF^{S_{1}}$ and $a$ is undefended by $S_{2}$ in $\AF^{S_{1}}$. This contradicts the assumption that $S_{2}$ is admissible in $\AF^{S_{1}}$.
	\end{proof}
\end{theorem}
By recursively applying the above theorem, we obtain the following corollary.
\begin{corollary}\label{cor:rec}
	Every non-empty admissible set $S$ can be written as $S=S_{1}\cup\ldots\cup S_{n}$ with pairwise disjoint $S_{i}$, $i=1,\ldots,n$, $S_{1}$ is an initial set of $\AF$ and every $S_{i}$, $i=2,\ldots,n$ is an initial set of $\AF^{S_{1}\cup\ldots \cup S_{i-1}}$. Furthermore, only non-empty admissible sets $S$ can be written in such a fashion.
\end{corollary}
\begin{example}\label{ex:core:const}
	Consider again the argumentation framework $\AF_0$ from Figure~\ref{fig:exex1} and recall that 
	\begin{align*}
		S_1 & = \{b,e,f,h,i\}
	\end{align*}
	is an admissible set of $\AF_{0}$ (and actually a preferred extension). $S_{1}$ can be written as
	\begin{align*}
		S_{1} & = \{h\}\cup\{f\}\cup\{e,i\}\cup\{b\}
	\end{align*}
	where
	\begin{itemize}
		\item $\{h\}$ is an initial set of $\AF_{0}$ (Figure~\ref{fig:exex1}),
		\item $\{f\}$ is an initial set of $\AF_{0}^{\{h\}}$ (Figure~\ref{fig:exex1:rev1}),
		\item $\{e,i\}$ is an initial set of $\AF_{0}^{\{h,f\}}$ (Figure~\ref{fig:exex1:rev2}), and
		\item $\{b\}$ is an initial set of $\AF_{0}^{\{e,i,h,f\}}$ (Figure~\ref{fig:exex1:rev3}).
	\end{itemize}
	\begin{figure}[t]
\begin{center}
\begin{subfigure}[b]{0.3\textwidth}
\begin{center}
         \begin{tikzpicture}[scale=0.6, every node/.style={scale=0.6},node distance=0.7cm]

	\node[args](a){$a$};
	\node[args, right=of a](b){$b$};
	\node[args, right=of b](c){$c$};
	\node[args, right=of c](d){$d$};
	\node[args, right=of d](e){$e$};
	\node[args, below=of a](f){$f$};

	\node[args, below=of d](i){$i$};
	\node[args, right=of i](j){$j$};
	
	\path(a) edge [->,loop,in=320,out=30,min distance=10mm] (a);
	\path(a) edge [->,bend left] (f);
	\path(f) edge [->,bend left] (a);
	\path(c) edge [->] (b);
	\path(d) edge [->] (c);
	\path(i) edge [->] (j);
	\path(j) edge [->] (e);
	\path(e) edge [->] (d);
	\path(d) edge [->] (i);
	\path(i) edge [->] (c);

\end{tikzpicture}
         \caption{$\AF_{0}^{\{h\}}$}
         \label{fig:exex1:rev1}
                  \end{center}
     \end{subfigure}~\\
     
     \medskip
     
\begin{subfigure}[b]{5cm}
\begin{center}
         \begin{tikzpicture}[scale=0.6, every node/.style={scale=0.6},node distance=0.7cm]

	\node[args](b){$b$};
	\node[args, right=of b](c){$c$};
	\node[args, right=of c](d){$d$};
	\node[args, right=of d](e){$e$};
	\node[args, below=of d](i){$i$};
	\node[args, right=of i](j){$j$};
	
	\path(c) edge [->] (b);
	\path(d) edge [->] (c);
	\path(i) edge [->] (j);
	\path(j) edge [->] (e);
	\path(e) edge [->] (d);
	\path(d) edge [->] (i);
	\path(i) edge [->] (c);

\end{tikzpicture}
         \caption{$\AF_{0}^{\{h,f\}}$}
         \label{fig:exex1:rev2}
         \end{center}
     \end{subfigure}
\begin{subfigure}[b]{3cm}
	\begin{center}
         \begin{tikzpicture}[scale=0.6, every node/.style={scale=0.6}]

	\node[args](b){$b$};

\end{tikzpicture}
         \caption{$\AF_{0}^{\{e,i,h,f\}}$}
         \label{fig:exex1:rev3}
         \end{center}
     \end{subfigure}

\end{center}
\caption{Reducts obtained from $\AF_0$ for the construction of $S_1 = \{b,e,f,h,i\}$.}
\label{fig:exex1:rev}
\end{figure}
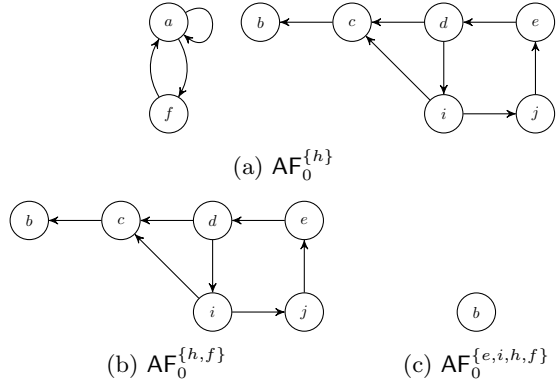
\end{example}
Note that a decomposition $S=S_{1}\cup\ldots\cup S_{n}$ of an admissible set $S$ from Corollary~\ref{cor:rec} is not necessarily uniquely determined. In the previous example, $S_{1}$ could also have been constructed by selecting, e.\,g., $\{f\}$ first.

Let us now discuss the wider significance of Theorem~\ref{th:elc1} and Corollary~\ref{cor:rec}. For that, recall the standard approach to compute and justify the (uniquely determined) grounded extension of an argumentation framework $\AFcomplete$ \cite{Dung:1995}, cf.\ also the discussion in \cite{Xu:2016}. Basically, the grounded extension $E$ of $\AF$ can be computed by selecting any non-attacked argument $a\in \arguments$, add it to $E$, remove $a$ and all arguments attacked by $a$ from $\AF$ (so move from $\AF$ to $\AF^{\{a\}}$), and continue the process until no further unattacked argument can be found. Observe the similarity of this procedure to the procedure indicated by Theorem~\ref{th:elc1}: in order to construct any admissible set $S$ of $\AF$, first select any initial set $S'$ of $\AF$, add it to $S$ (which is initially empty), remove $S'$ and all arguments attacked by $S'$ from $\AF$, and continue the process. Therefore, initial sets allow us to \emph{serialise} the construction of any admissible set into smaller steps, each of these steps solving a single conflict in the framework under consideration. Depending on how initial sets are selected at each step and how we end the process, we can also recover different semantics. Let us now formalise these ideas. For that, we first need two functions that define the selection mechanism of initial sets and the termination criterion.
\begin{definition}
	A \emph{state} $T$ is a tuple $T=(\AF,S)$ with $\AF\in\allAFs$ and $S\subseteq \allArgs$.
\end{definition}
\begin{definition}
	A \emph{selection function} $\alpha$ is any function $\alpha:2^{\allArgs}\times 2^{\allArgs}\times 2^{\allArgs}\rightarrow 2^{\allArgs}$ with $\alpha(X,Y,Z)\subseteq X\cup Y\cup Z$ for all $X,Y,Z\subseteq \allArgs$.
\end{definition}
We will apply a selection function $\alpha$ in the form $\alpha(\uelc(\AF),\ucelc(\AF),\celc(\AF))$ (for some $\AF)$, so $\alpha$ selects a subset of the initial sets as eligible to be selected in the construction process. We explicitly differentiate the different types of initial sets as parameters here as a technical convenience.
\begin{definition}
	A \emph{termination function} $\beta$ is any function $\beta:\allAFs\times 2^{\allArgs}\rightarrow \{0,1\}$.
\end{definition}
A termination function $\beta$ is used to indicate when a construction of an admissible set is finished (this will be the case if $\beta(\AF,S)=1$).

We will now define a transition system on states that makes use of a selection and a termination function to constrain the construction of admissible sets.
For some selection function $\alpha$, consider the following transition rule:
\begin{align}
	(\AF,S)\xrightarrow{S'\in\alpha(\uelc(\AF),\ucelc(\AF),\celc(\AF))} (\AF^{S'},S\cup S')\label{eq:trans}
\end{align}
If $(\AF',S')$ can be reached from $(\AF,S)$ via a finite number of steps (this includes no steps at all) with the above rule we write $(\AF,S)\leadsto^{\alpha}(\AF',S')$. If, in addition, the state $(\AF',S')$ also satisfies the termination criterion of $\beta$, i.\,e., $\beta(\AF',S)=1$, then we write $(\AF,S)\leadsto^{\alpha,\beta}(\AF',S')$.

Given concrete instances of $\alpha$ and $\beta$, let $\mathcal{E}^{\alpha,\beta}(\AF)$ be the set of all $S$ with $(\AF,\emptyset)\leadsto^{\alpha,\beta}(\AF',S)$ (for some $\AF'$). 
\begin{definition}
	A semantics $\sigma$ is \emph{serialisable} with a selection function $\alpha$ and a termination function $\beta$ if $\sigma(\AF)=\mathcal{E}^{\alpha,\beta}(\AF)$ for all $\AF$.
\end{definition}

A direct consequence of Corollary~\ref{cor:rec} is the following.
\begin{theorem}\label{th:char:adm}
	Admissible semantics\footnote{Although admissible sets are usually not regarded as a semantics, we can treat the function $\adm(\cdot)$ as such.} is serialisable with 
	\begin{align*}
		\alpha_{\adm}(X,Y,Z) & = X\cup Y\cup Z & \beta_{\adm}(\AF,S)=1
	\end{align*}
\end{theorem}
In other words, any admissible set can be constructed by not constraining the selection of initial sets at all ($\alpha_{\adm}$) and accepting every reachable state ($\beta_{\adm}$). We can also characterise most of the admissibility-based semantics from Section~\ref{sec:aa} through serialisation using specific selection and termination functions, as the following results show.

The following observation has been hinted at in \cite{Xu:2018}, but not formally proven. Using the notions of selection and termination functions, we can make the construction principle of complete extensions explicit.
\begin{theorem}\label{th:ser:comp}
	Complete semantics is serialisable with $\alpha_{\adm}$ and
	\begin{align*}
		\beta_{\co}(\AF,S)&=\left\{\begin{array}{ll}
							1 & \text{if $\uelc(\AF)=\emptyset$}\\
							0 & \text{otherwise}
							\end{array}\right.
	\end{align*}
\end{theorem}
Note that $\beta_{\co}$ only accepts those states, which cannot be extended by already defended arguments (which are those appearing in $\uelc(\AF)$). The above theorem also allows us to characterise complete extensions in a similar manner as admissible sets in Theorem~\ref{th:elc1}.
\begin{corollary}\label{cor:char:complete}
	Let  $\AFcomplete$ be an abstract argumentation framework and $S\subseteq \arguments$. $S$ is complete if and only if either
	\begin{itemize}
		\item $S=\emptyset$ and $\uelc(\AF)=\emptyset$ or
		\item $S=S_{1}\cup S_{2}$, $S_{1}\in\elc(\AF)$ and $S_{2}$ is complete in $\AF^{S_{1}}$.
	\end{itemize}
\end{corollary}
Grounded semantics has the same termination criterion as complete semantics, but constrains the selection of initial sets to those in $\uelc(\AF)$.
\begin{theorem}\label{th:grounded:ser}
	Grounded semantics is serialisable with 
	\begin{align*}
		\alpha_{\gr}(X,Y,Z) & = X
	\end{align*}
	and $\beta_{\co}$.
\end{theorem}
Note that $\alpha_{\gr}$ and $\beta_{\co}$ formalise the algorithm to compute the grounded extension sketched before. Therefore, the non-deterministic algorithm realised by the transition rule (\ref{eq:trans}) is a generalisation of this algorithm. Similarly as Corollary~\ref{cor:char:complete} we obtain the following characterisation of grounded semantics in terms of the reduct.
\begin{corollary}\label{cor:char:grounded}
	Let  $\AFcomplete$ be an abstract argumentation framework and $S\subseteq \arguments$. $S$ is grounded if and only if either
	\begin{itemize}
		\item $S=\emptyset$ and $\uelc(\AF)=\emptyset$ or
		\item $S=S_{1}\cup S_{2}$, $S_{1}\in\uelc(\AF)$ and $S_{2}$ is grounded in $\AF^{S_{1}}$.
	\end{itemize}
\end{corollary}
 
For stable semantics, we do not need to constrain the selection of initial sets but only ensure that all arguments are either included in or attacked by the constructed extension.
\begin{theorem}
	Stable semantics is serialisable with $\alpha_{\adm}$ and
	\begin{align*}
		\beta_{\st}(\AF,S)&=\left\{\begin{array}{ll}
							1 & \text{if $\AF=(\emptyset,\emptyset)$}\\
							0 & \text{otherwise}
							\end{array}\right.
	\end{align*}
\end{theorem}
As before, we obtain the following characterisation of stable semantics in terms of the reduct.
\begin{corollary}\label{cor:char:stable}
	Let  $\AFcomplete$ be an abstract argumentation framework and $S\subseteq \arguments$. $S$ is stable if and only if either
	\begin{itemize}
		\item $S=\emptyset$ and $\arguments=\emptyset$ or
		\item $S=S_{1}\cup S_{2}$, $S_{1}\in\elc(\AF)$ and $S_{2}$ is stable in $\AF^{S_{1}}$.
	\end{itemize}
\end{corollary}

For preferred semantics, we simply have to apply transitions exhaustively. Note that this result strengthens Proposition~3 from \cite{Xu:2018}.
\begin{theorem}
	Preferred semantics is serialisable with $\alpha_{\adm}$ and
	\begin{align*}
		\beta_{\pr}(\AF,S)&=\left\{\begin{array}{ll}
							1 & \text{if $\elc(\AF)=\emptyset$}\\
							0 & \text{otherwise}
							\end{array}\right.
	\end{align*}
\end{theorem}
\begin{corollary}\label{cor:char:preferred}
	Let  $\AFcomplete$ be an abstract argumentation framework and $S\subseteq \arguments$. $S$ is preferred if and only if either
	\begin{itemize}
		\item $S=\emptyset$ and $\elc(\AF)=\emptyset$ or
		\item $S=S_{1}\cup S_{2}$, $S_{1}\in\elc(\AF)$ and $S_{2}$ is preferred in $\AF^{S_{1}}$.
	\end{itemize}
\end{corollary}
Our final positive result is about strong admissibility, which follows quite easily from the construction of the grounded extension.
\begin{theorem}
	Strong admissibility semantics is serialisable with $\alpha_{\gr}$ and	$\beta_{\adm}$.
\end{theorem}
\begin{corollary}\label{cor:char:strongadm}
	Let  $\AFcomplete$ be an abstract argumentation framework and $S\subseteq \arguments$. $S$ is strongly admissible if and only if either
	\begin{itemize}
		\item $S=\emptyset$ or
		\item $S=S_{1}\cup S_{2}$, $S_{1}\in\uelc(\AF)$ and $S_{2}$ is strongly admissible in $\AF^{S_{1}}$.
	\end{itemize}
\end{corollary}
A related result to the above observation is given by Baumann et.\ al in \cite{Baumann:2016} (Definition 7 and Proposition 2). There, a strongly admissible extension $E$ is characterised through the existence of pairwise disjoint sets $A_1,\ldots,A_n$ such that $E=A_1\cup\ldots A_n$, $A_1$ is a set of unattacked arguments in $\AF$ and $A_1\cup\ldots\cup A_j$ defends $A_{j+1}$ for all $1\leq j \leq n-1$. Our construction above also implies the existence of these sets $A_1,\ldots,A_n$ with the same properties, but with the additional feature that all $A_i$, $i=1,\ldots,n$ are singleton sets (since unattacked initial sets are always singleton sets).  

Missing from our results so far are the ideal and semi-stable semantics. Both of them cannot be serialised. 
\begin{theorem}\label{th:ideal}
	Ideal semantics is not serialisable.
\end{theorem}
The proof of the above theorem is given by the following counterexample.
\begin{example}\label{ex:ideal}
		Consider the two argumentation frameworks $\AF_2$ and $\AF_{3}$ in Figure~\ref{fig:exexideal}.
		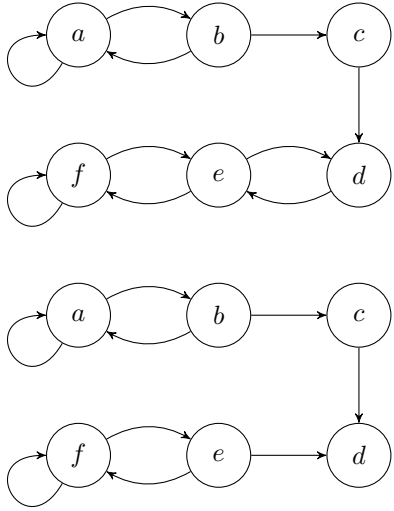
\begin{figure}[t]
\begin{center}
\begin{tikzpicture}

	\node[args](a){$a$};
	\node[args, right=of a](b){$b$};
	\node[args, right=of b](c){$c$};
	\node[args, below=of c](d){$d$};
	\node[args, below=of b](e){$e$};
	\node[args, below=of a](f){$f$};
	
	\path(a) edge [->,bend left] (b);
	\path(b) edge [->,bend left] (a);
	\path(b) edge [->] (c);
	\path(a) edge [->,loop,in=180,out=240,min distance=10mm] (a);

	\path(e) edge [->,bend left] (f);
	\path(f) edge [->,bend left] (e);	
	\path(c) edge [->] (d);
	\path(d) edge [->, bend left] (e);
	\path(e) edge [->, bend left] (d);
	\path(f) edge [->,loop,in=180,out=240,min distance=10mm] (f);

	\node[args, below=of f](ax){$a$};
	\node[args, right=of ax](bx){$b$};
	\node[args, right=of bx](cx){$c$};
	\node[args, below=of cx](dx){$d$};
	\node[args, below=of bx](ex){$e$};
	\node[args, below=of ax](fx){$f$};
	
	\path(ax) edge [->,bend left] (bx);
	\path(bx) edge [->,bend left] (ax);
	\path(bx) edge [->] (cx);
	\path(ax) edge [->,loop,in=180,out=240,min distance=10mm] (ax);

	\path(ex) edge [->,bend left] (fx);
	\path(fx) edge [->,bend left] (ex);	
	\path(cx) edge [->] (dx);
	\path(ex) edge [->] (dx);
	\path(fx) edge [->,loop,in=180,out=240,min distance=10mm] (fx);

\end{tikzpicture}
\end{center}
\caption{The argumentation frameworks $\AF_2$ (top) and $\AF_{3}$ (bottom) from Example~\ref{ex:ideal}.}
\label{fig:exexideal}
\end{figure}
		We have
		\begin{align*}
			\id(\AF_{2}) & = \{b\} & \id(\AF_{3}) & = \{b,e\}
		\end{align*}
		and
		\begin{align*}
			\uelc(\AF_{2}) & = \uelc(\AF_{3}) = \emptyset\\
			\ucelc(\AF_{2}) & = \ucelc(\AF_{3}) = \{\{b\},\{e\}\}\\
			\celc(\AF_{2}) & = \celc(\AF_{3}) = \emptyset
		\end{align*}		
		So no selection function $\alpha$ can distinguish these frameworks and should return either
		\begin{enumerate}
			\item $\alpha(\emptyset,\{\{b\},\{e\}\},\emptyset)=\emptyset$ or
			\item $\alpha(\emptyset,\{\{b\},\{e\}\},\emptyset)=\{\{b\},\{e\}\}$.
		\end{enumerate}
		Note also that $\alpha$ cannot return just one of the two initial sets as they cannot be distinguished. In case 1, the ideal extensions of both $\AF_{2}$ and $\AF_{3}$ cannot be constructed. So assume case 2 and select $\{e\}$ in the first transition. Observe that $AF_{2}^{\{e\}}=AF_{3}^{\{e\}}$, so further constructions will be identical. Since the ideal extensions of $\AF_{2}$ and $\AF_{3}$ differ and no $\beta$ can distinguish these cases, ideal semantics is not serialisable.
\end{example}
The above behaviour of ideal semantics is insofar surprising since the concept of unchallenged initial sets is closely related to the general idea of the ideal extension. Recall that the initial extension is the maximal admissible set contained in each preferred extension. This basically means that the arguments in the ideal extension are compatible with each admissible set and no admissible set attacks any argument in the ideal extension. On the other hand, an unchallenged initial set is likewise an undisputed admissible set, as there is no other initial set that attacks it. However, as the above example shows, unchallenged initial sets are not sufficient to characterise the ideal extension. We will discuss the relationship between unchallenged initial sets and the ideal extension a bit more below and in Section~\ref{sec:complexity}.

Likewise negative, but for somewhat different reasons, is the following result about semi-stable semantics. 
\begin{theorem}\label{th:ser:semi}
	Semi-stable semantics is not serialisable.
\end{theorem}
The reason that semi-stable semantics is not serialisable is that it needs some form of ``global'' view on candidate extensions to judge whether a set of arguments is indeed a semi-stable extension. We illustrate this in the next example (which also serves as the proof of Theorem~\ref{th:ser:semi}).
\begin{example}\label{ex:semi}
	Consider the three argumentation frameworks $\AF_{4}$, $\AF_{5}$, $\AF_{6}$ in Figure~\ref{fig:exexsemi}. Observe that 
		\begin{align*}
			\sst(\AF_{4})& =\{\{a,c\},\{b\}\}\\
			\sst(\AF_{5})& =\{\{b\}\}\\
			\sst(\AF_{6})& =\{\{a\},\{b\}\}
		\end{align*}
	However,
		\begin{align*}
			\uelc(\AF_{4})&=\uelc(\AF_{5})=\uelc(\AF_{6})=\emptyset\\
			\ucelc(\AF_{4})&=\ucelc(\AF_{5})=\ucelc(\AF_{6})=\emptyset\\
			\celc(\AF_{4}) &=\celc(\AF_{5})=\celc(\AF_{6})=\{\{a\},\{b\}\}
		\end{align*}
	First, no selection function $\alpha$ can distinguish these frameworks (as it only operates on the sets $\uelc(\cdot)$, $\ucelc(\cdot)$, and $\celc(\cdot)$) and would either return
	\begin{enumerate}
		\item $\alpha(\emptyset,\emptyset,\{\{a\},\{b\}\})=\emptyset$,
		\item $\alpha(\emptyset,\emptyset,\{\{a\},\{b\}\})=\{\{a\}\}$,
		\item $\alpha(\emptyset,\emptyset,\{\{a\},\{b\}\})=\{\{b\}\}$, or
		\item $\alpha(\emptyset,\emptyset,\{\{a\},\{b\}\})=\{\{a\},\{b\}\}$.
	\end{enumerate}
	In cases 1--3, not all semi-stable extensions can be constructed for, e.\,g., $\AF_{4}$. For case 4\ and $\AF_{5}$, a valid transition would then produce the state $T_{1}=((\{c\},\{(c,c)\}),\{a\})$ from which no further transition is possible. A $\beta$ function for semi-stable semantics should now determine that $T_{1}$ is not a terminating state (since $\{a\}$ is not a semi-stable extension of $\AF_{5}$). However, the same state $T_{1}$ can also be reached for $\AF_{6}$, but here $\{a\}$ is a semi-stable extension. Since $\beta$ cannot distinguish these two scenarios at $T_{1}$, there is no such $\beta$.
	\begin{figure}[t]
\begin{center}
\begin{tikzpicture}[scale=1, every node/.style={scale=1},node distance=0.7cm]

	\node[args](a){$a$};
	\node[args, right=of a](b){$b$};
	\node[args, right=of b](c){$c$};
	
	\path(a) edge [->,bend left] (b);
	\path(b) edge [->,bend left] (a);
	\path(b) edge [->] (c);
	
	\node[args, below= of a](ax){$a$};
	\node[args, right=of ax](bx){$b$};
	\node[args, right=of bx](cx){$c$};
	
	\path(ax) edge [->,bend left] (bx);
	\path(bx) edge [->,bend left] (ax);
	\path(bx) edge [->] (cx);
	\path(cx) edge [->,loop,in=0,out=60,min distance=10mm] (cx);
	
	\node[args, below= of ax](axx){$a$};
	\node[args, right=of axx](bxx){$b$};
	\node[args, right=of bxx](cxx){$c$};
	
	\path(axx) edge [->,bend left] (bxx);
	\path(bxx) edge [->,bend left] (axx);
	\path(cxx) edge [->,loop,in=0,out=60,min distance=10mm] (cxx);

\end{tikzpicture}
\end{center}
\caption{The argumentation frameworks $\AF_4$ (top), $\AF_5$ (middle), and $\AF_{6}$ (bottom) from Example~\ref{ex:semi}.}
\label{fig:exexsemi}
\end{figure}
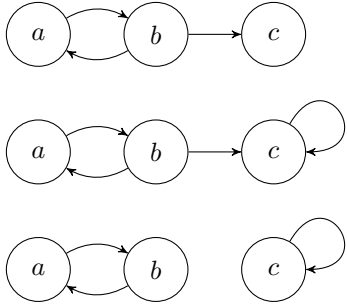
\end{example}
The same argument from above can also be used to show that \emph{eager semantics} is not serialisable. The \emph{eager extension} is the maximal (wrt.\ set inclusion) admissible set contained in every semi-stable extension, see e.\,g.\ \cite{Baroni:2018a}. In Example~\ref{ex:semi}, the eager extension of $\AF_{5}$ is $\{b\}$ while it is $\emptyset$ for $\AF_{4}$ and $\AF_{6}$. No pair $(\alpha,\beta)$ can be defined to distinguish these cases as well.

The results of this section show that many admissibility-based semantics can be characterised through the notion of initial sets and a simple non-deterministic algorithm based on selecting initial sets at each step. This brings a new perspective on the rationality of admissibility-based semantics, as their basic construction principles are made explicit via an operational mechanism. This is similar in spirit to the purpose of \emph{discussion games} \cite{Caminada:2018x} which model acceptability problems of individual arguments as an operational mechanism as well (here a dialogue between a proponent and an opponent). However, here we focused on the construction of whole extensions and not on acceptability problems of individual arguments.

The notion of serialisability also allows to define completely new semantics by defining only a selection and a termination function. For example, a straightforward idea for that would be the selection function $\alpha_{0}$ defined via 
\begin{align*}
	\alpha_{0}(X,Y,Z) & = X\cup Y
\end{align*}
and the termination function $\beta_{0}$ defined via
\begin{align*}
	\beta_{0}(\AF,S) & = \left\{\begin{array}{ll}
							1 & \text{if $\uelc(\AF)\cup \ucelc(\AF)=\emptyset$}\\
							0 & \text{otherwise}
							\end{array}\right.
\end{align*}
which amounts to exhaustively adding unattacked and unchallenged initial sets. This yields a semantics that is more skeptical than the preferred semantics but less skeptical than the ideal semantics as the following result shows.
\begin{theorem}\label{th:initial:ideal}
	For every $E$ with $(\AF,\emptyset)\leadsto^{\alpha_{0},\beta_{0}}(\AF',E)$
	\begin{enumerate}
		\item $E\subseteq E'$ for some preferred extension $E'$ and
		\item $E_{\id}\subseteq E$ for the ideal extension $E_{\id}$.
	\end{enumerate}
\end{theorem}
Consider the following examples showing the difference between the above semantics and ideal semantics.
\begin{example}
	Consider again $\AF_2$ from Example~\ref{ex:ideal} and depicted in Figure~\ref{fig:exexideal}. There are two extensions $E_1$ and $E_2$ wrt.\ to the serialisable semantics defined by $\alpha_0$ and $\beta_0$:
	\begin{align*}
		E_1 & = \{b\}	& E_2 & = \{b,e\}
	\end{align*}
	where the extension $E_2$ can be constructed by first selecting the initial set $\{e\}$ (which is unchallenged in $\AF_2$) and then $\{b\}$. 
\end{example}
\begin{example}\label{ex:exalpha0beta0}
	Consider $\AF_7$ depicted in Figure~\ref{fig:exalpha0beta0}. There are four preferred extensions $E_1$, $E_2$, $E_3$, and $E_4$ in $\AF_7$ defined via
		\begin{align*}
		 E_1 & = \{a,e\}		& E_2 & = \{a,d,f\} &
		 E_3 & = \{b,e\}		& E_4 & = \{b,d,f\}
		\end{align*}
		and the ideal extension $E_{id}$ is empty:
		\begin{align*}
			E_{id}	& = \emptyset
		\end{align*}
		However, there is one extension $E_5$ wrt.\ to the serialisable semantics defined by $\alpha_0$ and $\beta_0$:
		\begin{align*}
			E_5 & = \{d,f\}
		\end{align*}
		The reason for that is that both $\{d\}$ and $\{f\}$ are unchallenged initial sets in $\AF_7$ (and once one is selected the other becomes an unattacked initial set and can be selected as well). 
	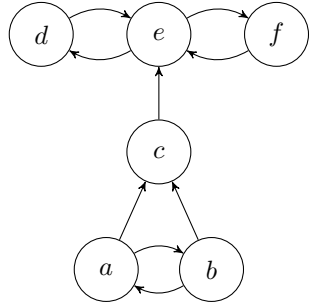
\begin{figure}[t]
\begin{center}
\begin{tikzpicture}[scale=1, every node/.style={scale=1},node distance=0.7cm]

	\node[args](d){$d$};
	\node[args, right=of d](e){$e$};
	\node[args, right=of e](f){$f$};
	\node[args, below=of e](c){$c$};
	\node[args, below=of c,xshift=-0.7cm](a){$a$};
	\node[args, below=of c,xshift=0.7cm](b){$b$};
	
	\path(a) edge [->,bend left] (b);
	\path(b) edge [->,bend left] (a);
	\path(b) edge [->] (c);
	\path(a) edge [->] (c);
	\path(c) edge [->] (e);
	
	\path(e) edge [->,bend left] (d);
	\path(d) edge [->,bend left] (e);
	\path(e) edge [->,bend left] (f);
	\path(f) edge [->,bend left] (e);

\end{tikzpicture}
\end{center}
\caption{The argumentation framework $\AF_7$ from Example~\ref{ex:exalpha0beta0}.}
\label{fig:exalpha0beta0}
\end{figure}	
\end{example}
For future work, we intend to investigate the above and further possibilities for serialisable semantics in more detail.

\section{Computational Complexity}\label{sec:complexity}
In order to round up our investigation of initial sets, we now analyse them wrt.\ computational complexity. We assume familiarity with basic concepts of computational complexity and basic complexity classes such as $\textsf{P}$, $\textsf{NP}$, $\textsf{coNP}$, see \cite{Papadimitriou:1994a} for an introduction. We also require knowledge of the ``non-standard'' classes \textsf{DP} (and its complement \textsf{coDP}), $\textsf{P}^{\textsf{NP}}$, and $\textsf{P}^{\textsf{NP}}_{\parallel}$. The class \textsf{DP} is the class of decision problems that are a conjunction of a problem in \textsf{NP} and a problem in \textsf{coNP}, i.\,e., in language notation $\textsf{DP}=\{L_{1}\cap L_{2}\mid L_{1}\in \textsf{NP},L_{2}\in \textsf{coNP}\}$.
The class $\textsf{P}^{\textsf{NP}}$ is the class of decision problems that can be solved by a deterministic polynomial-time algorithm that can make polynomially many \emph{adaptive} queries\footnote{A query is adaptive if it may depend on a previous query; queries are non-adaptive if they can be posed in any order or in parallel} to an \textsf{NP}-oracle.
 The class $\textsf{P}^{\textsf{NP}}_{\parallel}$ \cite{Eiter:1997} is the class of decision problems that can be solved by a deterministic polynomial-time algorithm that can make polynomially many \emph{non-adaptive} queries to an \textsf{NP}-oracle. Note that $\textsf{P}^{\textsf{NP}}_{\parallel}$ is sometimes denoted by $\Theta^{P}_{2}$ and is equal to $\textsf{P}^{\textsf{NP}[log]}$, i.\,e., the class of decision problems solvable by deterministic polynomial-time algorithm that can make logarithmically many \emph{adaptive} \textsf{NP}-oracle calls \cite{Papadimitriou:1994a}. Observe also that $\textsf{DP}\subseteq\textsf{P}^{\textsf{NP}}_{\parallel}$ and $\textsf{coDP}\subseteq\textsf{P}^{\textsf{NP}}_{\parallel}$ as well as $\textsf{P}^{\textsf{NP}}_{\parallel} \subseteq \textsf{P}^{\textsf{NP}}$.

We consider the following computational tasks for $\sigma\in\{\elc,\uelc,\ucelc,\celc\}$, cf.\ \cite{Dvorak:2018}:
\begin{tabbing}
	\,\,\=$\textit{Ver}_{\sigma}$ \qquad \= Given $\AFcomplete$ and $S\subseteq\arguments$,\\\>\> decide whether $S\in \sigma(\AF)$.\\
	\>$\textit{Exists}_{\sigma}$ 	\> Given $\AFcomplete$,\\\>\>decide whether $\sigma(\AF)\neq \emptyset$.\\
	\>$\textit{Unique}_{\sigma}$	\> Given $\AFcomplete$,\\\>\>decide whether $|\sigma(\AF)|=1$.\\ 
	\>$\textit{Cred}_{\sigma}$	\> Given $\AFcomplete$ and $a\in\arguments$,\\\>\>decide whether there is $S\in \sigma(\AF)$ with $a\in S$.\\ 
	\>$\textit{Skept}_{\sigma}$	\> Given $\AFcomplete$ and $a\in\arguments$,\\\>\>decide whether for all $S\in \sigma(\AF)$, $a\in S$.
\end{tabbing}
Note that we do not consider the problems $\textit{Exists}_{\sigma}^{\neg \emptyset}$ \cite{Dvorak:2018} (which asks whether there is a non-empty (initial) set) as these are equivalent to $\textit{Exists}_{\sigma}$ due to the non-emptiness of all types of initial sets. 

Table~\ref{tbl:comp} summarises our results on the complexity of the above tasks, all proofs can be found in the appendix. 

\begin{table}\begin{center}\setlength{\tabcolsep}{5pt}
	\begin{tabular}{| l | c| c|c|c|}
		\hline
		$\sigma$							&	$\elc$		& $\uelc$ & $\ucelc$ & $\celc$ \\
		\hline\hline
		$\textit{Ver}_{\sigma}$				&	in \textsf{P}	& in \textsf{P} & \textsf{coNP}-c&\textsf{NP}-c \\[0.6mm]
		$\textit{Exists}_{\sigma}$				&	\textsf{NP}-c	& in \textsf{P} & $\textsf{P}^{\textsf{NP}}_{\parallel}$-c& \textsf{NP}-c\\[0.6mm]
		$\textit{Unique}_{\sigma}$				&	\textsf{DP}-c& in \textsf{P} & in $\textsf{P}^{\textsf{NP}}_{\parallel}$, \textsf{DP}-h & trivial \\[0.6mm]
		$\textit{Cred}_{\sigma}$				&	\textsf{NP}-c	& in \textsf{P} & $\textsf{P}^{\textsf{NP}}_{\parallel}$-c& \textsf{NP}-c\\[0.6mm]
		$\textit{Skept}_{\sigma}$				&	\textsf{coNP}-c& in \textsf{P} & $\textsf{P}^{\textsf{NP}}_{\parallel}$-c &\textsf{coNP}-c 	\\[0.6mm]\hline
	\end{tabular}
	\end{center}
	\caption{Complexity of computational tasks related to initial sets. An attached ``-h'' refers to hardness for the class and an attached ``-c'' refers to completeness for the class. All hardness results are wrt.\ polynomial reductions.}
	\label{tbl:comp}
\end{table}

The results for $\elc$ mirror the results on admissible sets \cite{Dvorak:2018}, with a few exceptions. While the problem of deciding whether a set $S$ is admissible can be solved in logarithmic space with polynomial time \cite{Dvorak:2018}, we only showed that the problem of deciding whether a set $S$ is an initial set can be solved in polynomial time. It is unlikely (though a formal proof is missing at the moment) that we can strengthen these results in the same way as for admissible sets, since the minimality condition of an initial set suggest that subsets (of linear size) have to be constructed in an algorithm. Moreover, while the problems $\textit{Exists}$ and $\textit{Skept}$ are trivial for admissible sets \cite{Dvorak:2018} (since the empty set is always admissible), they are intractable for initial sets. The problem $\textit{Unique}_{\elc}$ is \textsf{DP}-complete (it is \textsf{coNP}-complete for admissible sets) since existence of initial sets is not guaranteed.

We get some different complexity characterisations for the different types of initial sets. All tasks for unattacked initial sets are tractable, as only unattacked arguments have to be considered. All tasks become harder (under standard complexity-theoretic assumptions) when only unchallenged initial sets are considered. In particular, $\textit{Ver}_{\sigma}$ is \textsf{coNP}-complete as it has to be verified that there is no other initial set attacking the set that is to be verified. Moreover, Theorem~\ref{th:initial:ideal} already showed that there is some conceptual relation between unchallenged initial sets and the ideal extension.
This is strengthened by our observation of the computational complexity of the other tasks pertaining to unchallenged initial sets, as all non-trivial tasks on ideal semantics are $\textsf{P}^{\textsf{NP}}_{\parallel}$-complete (under \emph{randomised reductions}) \cite{Dunne:2009}. We showed $\textsf{P}^{\textsf{NP}}_{\parallel}$-completeness (under \emph{polynomial reductions}) for most of those tasks as well, with the exception being the problem $\textit{Unique}_{\ucelc}$, where we only showed $\textsf{DP}$-hardness and a $\textsf{P}^{\textsf{NP}}_{\parallel}$-hardness proof remains an open problem.

The complexity of the tasks for challenged initial sets are similar as in the general case, with two exceptions. Verification of challenged initial sets is intractable as another initial set has to found that attacks the set under consideration. Moreover, $\textit{Unique}_{\celc}$ has the trivial answer \textsc{NO} for all instances as the existence of one challenged initial set $S_{1}$ implies the existence of another challenged initial set $S_{2}$ that attacks (and is attacked by) $S_{1}$.
%
%
%
%
%
\section{Discussion}\label{sec:summary}
In this paper, we revisited the notion of initial sets, i.\,e., non-empty minimal admissible sets. We investigated their general properties and used them as basic building blocks to construct any admissible set. We have characterised many admissibility-based semantics via this approach and concluded our analysis with some notes on computational complexity.

Initial sets allow us to concisely explain why a certain argument can be accepted (e.\,g., whether it is contained in a preferred extension). We can deconstruct an extension via the characterisation of Corollary~\ref{cor:rec}, justify the inclusion of initial sets along this characterisation---e.\,g., by pointing to the conflicts that had to be solved---, and arrive step by step at the argument in question. 

A recent paper that addresses a similar topic as we do is \cite{Baumann:2021b}. There, Baumann and Ulbricht introduce \emph{explanation schemes} as a way to explain the construction of extensions wrt.\ complete, admissible, and strongly admissible semantics. Let us consider the case of complete semantics. Given $\AFcomplete$, the construction follows basically three steps\footnote{Although an abbreviated two-step procedure is also discussed.}: 1.) Determine the grounded extension $E_0$ of $\AF$, 2.) select a conflict-free set $E_1$ from the set of arguments appearing in some even-length cycle in the reduct $\AF^{E_0}$, and 3.) determine the grounded extension $E_2$ of the reduct $\AF^{E_0\cup E_1}$. If $E=E_0\cup E_1 \cup E_2$ defends $E_1$ then $E$ is a complete extension (and every complete extension can be decomposed in such a way). The overall construction is similar to our approach, in particular, it consists of a series of steps where we select a set of arguments and move to the reduct of the framework wrt.\ the arguments accumulated so far. However, our approach provides a more fine-grained way to construct extensions. In each step, we solve a single issue by selecting a single initial set. Step 2 of Baumann and Ulbricht's approach possibly solves a series of different conflicts all at once. This may actually diffuse the goal to provide an explanation why a certain extension is constructed as it is, as a selection of a conflict-free set of arguments from all even cycles may not clearly show, which conflicts are actually resolved. Furthermore, we do not need to explicitly use the notion of grounded semantics (and the general possibility to include defended arguments) in our construction, as it arises naturally through selecting (unattacked) initial sets and moving to the reduct immediately. 

As a by-product of our work, we introduced a new \emph{principle} \cite{vanderTorre:2018} for abstract argumentation semantics: \emph{serialisability}. For future work, we aim at investigated relationships of this new principle to other existing principles listed in \cite{vanderTorre:2018}. 
Another avenue for future work to investigate whether our characterisations of admissibility-based semantics can be exploited for algorithmic purposes \cite{Cerutti:2018a}. It is clear, that there is no obvious advantage in terms of computational complexity by computing (for example) a preferred extension via our transition system as it involves the computation of initial sets at each step (which is an intractable problem as $\textit{Exists}_{\elc}$ is already \textsf{NP}-complete). However, our characterisation of initial sets through strongly connected components (see Section~\ref{sec:cores}) could be exploited to devise a parallel algorithm---see also \cite{Cerutti:2015}---, as initial sets can be calculated independently of each other in each strongly connected component.

%

\section*{Appendix A. Proofs of technical results}
\setcounter{proposition}{0}
\setcounter{theorem}{1}
\setcounter{corollary}{0}
\begin{proposition}\label{prop:scc1}
	If $S$ is an initial set of $\AF$ then there is $\AF'=(\arguments',\attacks')\in\SCC(\AF)$ s.t.\ $S\subseteq \arguments'$.
	\begin{proof}
		If $S$ is a singleton the claim is trivially true. So let $S$ be an initial set of $\AF$ with $|S|>1$. Assume there are two different $\AF'=(\arguments',\attacks')\in\SCC(\AF)$ and $\AF''=(\arguments'',\attacks'')\in\SCC(\AF)$ with $S\cap \arguments'\neq \emptyset$ and $S\cap \arguments''\neq \emptyset$ and also $S=(S\cap \arguments')\cup (S\cap\arguments'')$ (the following proof generalises easily if $S$ is spanned across more than two SCCs). If $(S\cap \arguments')^{-}\subseteq (S\cap\arguments')^{+}$ then $S\cap\arguments'$ is admissible, contradicting the fact that $S$ is an initial set. So there is at least one $a\in (S\cap \arguments')^{-}$ with $a\in (S\cap \arguments'')^{+}$ since the complete set $S$ is admissible. With the same reasoning there must be a $b\in (S\cap \arguments'')^{-}$ with $b\in (S\cap \arguments')^{+}$. Then there is a closed circuit: from $a$ there is an edge to an element $c$ in $(S\cap \arguments')$. Since $(S\cap \arguments')$ is part of an \SCC, there is a path from $c$ to any other argument in $(S\cap \arguments')$, in particular, also to an attacker $d$ of $b$. From $d$ we can go to $b$ and then to an element in $(S\cap \arguments'')$. Again through the \SCC\ of $(S\cap \arguments'')$ we can reach $a$. This contradicts the assumption, so $S$ is contained in a single \SCC.			
	\end{proof}
\end{proposition}
\begin{proposition}\label{prop:scc2}
	$S$ is an initial set of $\AF$ if and only if $S$ is an initial set of $\SCC(S)=(\arguments',\attacks')$ and $S^{-}\subseteq \arguments'$
	\begin{proof}
		Let $S$ be an initial set of $\AF$. As $S$ is admissible, all arguments $a\in E^{-}$ are attacked by some $b\in E$. Proposition~\ref{prop:scc1} already established that $S\subseteq \arguments'$. Since each $a\in S^{-}$ attacks and is attacked by $S$, $S^{-}\subseteq \arguments'$ as well. It follows that $S$ is an initial set of $\SCC(S)$ as well as $S\cup S^{-} \subseteq \arguments'$ and verifying whether a set is initial only needs to consider the relationships of those arguments. This also proves the other direction.
	\end{proof}
\end{proposition}
\begin{proposition}
	Let $S\in\elc(\AF)$ and $\SCC(S)=(\arguments',\attacks')$.
	\begin{enumerate}
		\item If $S$ is unattacked then $|\arguments'|=1$.
		\item If $S$ is challenged or unchallenged then $|\arguments'|>1$.
		\item If $S$ is challenged and $S'\in\conflicts(S,\AF)$ then $\SCC(S)=\SCC(S')$.
	\end{enumerate}
	\begin{proof}
		Let $S\in\elc(\AF)$ and $\SCC(S)=(\arguments',\attacks')$.
		\begin{enumerate}
			\item Since $S=\{a\}$ is not attacked there is no other argument $b$ with a path to $a$. It follows directly $|\arguments'|=1$.
			\item As $S$ is admissible and there is at least one $b$ that attacks $S$, one argument in $S$ must attack $b$. It follows $S\cup\{b\}\in\SCC(S)$ and therefore $|\arguments'|>1$.
			\item Let $S'\in\conflicts(S,\AF)$. As $S\attacks S'$ and $S'\attacks S$ as well as both $S$ and $S'$ are completely in one \SCC, it follows $\SCC(S)=\SCC(S')$.\qedhere
		\end{enumerate}
	\end{proof}
\end{proposition}
\begin{proposition}
	Let  $\AFcomplete$ be an abstract argumentation framework and $S,S'\in \elc(\AF)$ with $S\neq S'$.
	\begin{enumerate}
		\item If $S'\in \uelc(\AF)$ then $S'\in \uelc(\AF^{S})$
		\item If $S'\in \conflicts(S,\AF)$ then $S'\notin \elc(\AF^{S})$
		\item If $S'\notin\conflicts(S,\AF)$ then $S'\cap \bigcup\elc(\AF^{S})\neq \emptyset$
	\end{enumerate}
	\begin{proof}
		Let  $\AFcomplete$ be an abstract argumentation framework and $S,S'\in \elc(\AF)$ with $S\neq S'$
		\begin{enumerate}
			\item Let $S'\in \uelc(\AF)$, so $S=\{a\}$. As $a$ is not attacked in $\AF$, it is also not attacked by $S$ and it follows $a\in \arguments'$ for $\AF^{S}=(\arguments',\attacks')$. It also follows that $a$ is not attacked in $\AF^{S}$ and therefore $\{a\}\in\uelc(\AF^{S})$.
			\item If $S'\in \conflicts(S,\AF)$ then there is $a\in S$ and $b\in S'$ with $a\attacks b$. It follows $b\notin \arguments'$ for $\AF^{S}=(\arguments',\attacks')$. So  $S'\notin \elc(\AF^{S})$.
			\item As $S'\notin\conflicts(S,\AF)$ it follows that $S\cup S'$ is conflict-free and therefore $S'\subseteq \arguments'$ for $\AF^{S}=(\arguments',\attacks')$. Furthermore, since $S'$ is admissible in $\AF$, $S'\cap \arguments'$ remains admissible in $\AF^{S}$. By definition, it follows that there is an initial set $S''$ of  $\AF^{S}$ with $S''\subseteq S'$, proving the claim.\qedhere
		\end{enumerate}
	\end{proof}
\end{proposition}
\begin{corollary}\label{cor:rec}
Every non-empty admissible set $S$ can be written as $S=S_{1}\cup\ldots\cup S_{n}$ with pairwise disjoint $S_{i}$, $i=1,\ldots,n$, $S_{1}$ is an initial set of $\AF$ and every $S_{i}$, $i=2,\ldots,n$ is an initial set of $\AF^{S_{1}\cup\ldots \cup S_{i-1}}$. Furthermore, only non-empty admissible sets $S$ can be written in such a fashion.
	\begin{proof}
		This follows by iterative application of Theorem~\ref{th:elc1}.
	\end{proof}
\end{corollary}
\begin{theorem}\label{th:char:adm}
	Admissible semantics is serialisable with 
	\begin{align*}
		\alpha_{\adm}(X,Y,Z) & = X\cup Y \cup Z & \beta_{\adm}(\AF,S)=1
	\end{align*}
	\begin{proof}
		Follows from Corollary~\ref{cor:rec}.
	\end{proof}
\end{theorem}
\begin{theorem}
	Complete semantics is serialisable with $\alpha_{\adm}$ and
	\begin{align*}
		\beta_{\co}(\AF,S)&=\left\{\begin{array}{ll}
							1 & \text{if $\uelc(\AF)=\emptyset$}\\
							0 & \text{otherwise}
							\end{array}\right.
	\end{align*}
	\begin{proof}
		We have to show that $E$ is a complete extension if and only if $(\AF,\emptyset)\leadsto^{\alpha_{\adm},\beta_{\co}}(\AF',E)$ for some $\AF'$. Let $\AFcomplete$.
		\begin{itemize}
			\item ``$\Rightarrow$'':\\
			Let $E$ be a complete extension. By Corollary~\ref{cor:rec} and the fact that $\alpha_{\adm}$ does not constrain the selection of initial sets, it is clear that there is $\AF'$ with $(\AF,\emptyset)\leadsto^{\alpha}(\AF',E)$. It remains to show that $\beta_{\co}(\AF',E)=1$. As $E$ is complete, there is no argument $a\in\arguments$ s.t.\ all attackers of $a$ (in $\AF$) are contained in $E$. This is equivalent to stating that there is no unattacked argument $a$ in $\AF^{E}$ and therefore $\uelc(\AF^{E})=\emptyset$. As $\AF^{E}=\AF'$ the claim follows.
			\item ``$\Leftarrow$'':\\
			Let $(\AF,\emptyset)\leadsto^{\alpha_{\adm},\beta_{\co}}(\AF',E)$ for some $\AF'=(\arguments',\attacks')$. By Corollary~\ref{cor:rec}, $E$ is admissible. Since $\uelc(\AF')=\emptyset$ there is no argument $a\in \arguments'$ that is defended by $E$. Therefore, $E$ is complete.\qedhere
		\end{itemize}
	\end{proof}
\end{theorem}
\begin{corollary}
	Let  $\AFcomplete$ be an abstract argumentation framework and $S\subseteq \arguments$. $S$ is complete if and only if either
	\begin{itemize}
		\item $S=\emptyset$ and $\uelc(\AF)=\emptyset$ or
		\item $S=S_{1}\cup S_{2}$, $S_{1}\in\elc(\AF)$ and $S_{2}$ is complete in $\AF^{S_{1}}$.
	\end{itemize}
	\begin{proof}
		If $\uelc(\AF)=\emptyset$ then $S=\emptyset$ is obviously a complete extension. Let $S$ be any non-empty complete extension. Due to Theorem~\ref{th:elc1} there are sets $S_1$, $S_2$ with $S=S_{1}\cup S_{2}$, $S_{1}\in\elc(\AF)$ and $S_{2}$ is admissible in $\AF^{S_{1}}$. Assume $S_2$ is not complete in $\AF^{S_{1}}$. Then there is an argument $a$ in $\AF^{S_{1}}$ that is defended by $S_2$ but $a\notin S_2$. But then $a\notin S_1$ and $a\notin S_1^+$ (otherwise it would not be in $\AF^{S_{1}}$) and $S$ defends $a$ as well in $\AF$. This contradicts the assumption that $S$ is complete. So $S_2$ is complete in $\AF^{S_{1}}$.
			
		For the other direction, let $S$ be any set with $S=S_{1}\cup S_{2}$, $S_{1}\in\elc(\AF)$ and $S_{2}$ is complete in $\AF^{S_{1}}$. Assume that $S$ is not complete in $\AF$. Then there is an argument $a$ that is defended by $S$ but $a\notin S$. Since $a$ is not contained in $S_1$ nor attacked by it, it follows that $a$ is contained in $\AF^{S_{1}}$. Let $T$ be the attackers of $a$ in $\AF$. Since $S$ attacks all arguments in $T$ let $T_1\subseteq T$ be those arguments in $T$ attacked by some $b\in S_1$ and $T_2=T\setminus T_1$. It follows that no argument of $T_1$ is contained in $\AF^{S_{1}}$ but all arguments in $T_2$ are contained in $\AF^{S_{1}}$ and are necessarily attacked by $S_2$. It follows that $a$ is defended by $S_2$ in $\AF^{S_{1}}$, contradicting the fact that $S_2$ is complete in $\AF^{S_{1}}$. It follows that $S$ is indeed complete.
	\end{proof}
\end{corollary}
\begin{theorem}
	Grounded semantics is serialisable with 
	\begin{align*}
		\alpha_{\gr}(X,Y,Z) & = X
	\end{align*}
	and $\beta_{\co}$.
	\begin{proof}
		Let $E_{gr}$ be the grounded extension of $\AFcomplete$ and let $E_k$ be such that 
			\begin{align*}
				(\AF,\emptyset)\rightarrow^{\alpha_{\gr}}(\AF_1,E_1)	\rightarrow^{\alpha_{\gr}}\ldots \rightarrow^{\alpha_{\gr}} (\AF_k,E_k)	
			\end{align*}
			so that $\beta_{\co}(\AF_k,E_k)=1$. We first show $E_i\subseteq E_{gr}$, for $i=1,\ldots,k$, by induction on $i$.
			\begin{itemize}
				\item $i=1$: By definition of $\alpha_{\gr}$ we have $E_1=\{a\}$ for some unattacked argument $a\in\arguments$. Since $E_{gr}$ is complete, it contains all unattacked arguments of $\arguments$, therefore $E_1\subseteq E_{gr}$.
				\item $i > 1$: By definition of $\alpha_{\gr}$ we have $E_i=\{a\}\cup E_{i-1}$ for some unattacked argument $a\in\arguments$ in $\AF_i$. Since $a$ is unattacked in $\AF_i$, all attackers of $a$ in $\AF$ must be attacked by $E_{i-1}$. By assumption, $E_{i-1}\subseteq E_{gr}$ and $a$ is defended by $E_{gr}$ as well. Since $E_{gr}$ is complete it follows $E_{i}\subseteq E_{gr}$.
			\end{itemize}
			Since Theorem~\ref{th:ser:comp} already established that $E_k$ is complete and $E_{gr}$ is the smallest complete extension, from $E_k\subseteq E_{gr}$ is follows $E_k=E_{gr}$ and therefore the claim.		
		\end{proof}
\end{theorem}
\begin{corollary}
	Let  $\AFcomplete$ be an abstract argumentation framework and $S\subseteq \arguments$. $S$ is grounded if and only if either
	\begin{itemize}
		\item $S=\emptyset$ and $\uelc(\AF)=\emptyset$ or
		\item $S=S_{1}\cup S_{2}$, $S_{1}\in\uelc(\AF)$ and $S_{2}$ is grounded in $\AF^{S_{1}}$.
	\end{itemize}
	\begin{proof}
		If $\uelc(\AF)=\emptyset$ then $S=\emptyset$ is obviously the grounded extension. Assume $S$ is the non-empty grounded extension of $\AF$. Then there must be an argument $a\in S$ that is not attacked in $\AF$. It follows $\{a\}\in\uelc(\AF)$. Let $S'$ be the grounded extension of $\AF^{\{a\}}$. Then $S'\cup\{a\}$ is complete in $\AF$:
		\begin{enumerate}
			\item $S'\cup\{a\}$ is admissible due to Theorem~\ref{th:elc1}.
			\item $a$ is defended by $S'\cup\{a\}$ as it is not attacked.
			\item every $b\in S'$ is defended by $S'\cup\{a\}$ as all attackers of $b$ are either attacked by $a$ (and therefore not in the reduct $\AF^{\{a\}}$) or some argument in $S'$ (since $S'$ is grounded in $\AF^{\{a\}}$).
		\end{enumerate}
		Assume there is a proper subset $S''\subset S'\cup\{a\}$ that is complete. Since $a$ is not attacked, $a\in S''$. It can easily be seen that $S'\cap S''$ would be complete in $\AF^{\{a\}}$ and $S'\cap S''\subset S'$, contradicting the fact that $S'$ is the grounded extension of $\AF^{\{a\}}$. It follows that $S=S'\cup\{a\}$.
			
		The other direction is analogous.
	\end{proof}
\end{corollary}
\begin{theorem}
	Stable semantics is serialisable with $\alpha_{\adm}$ and
	\begin{align*}
		\beta_{\st}(\AF,S)&=\left\{\begin{array}{ll}
							1 & \text{if $\AF=(\emptyset,\emptyset)$}\\
							0 & \text{otherwise}
							\end{array}\right.
	\end{align*}
	\begin{proof}
		Let $S$ be a stable extension. By Theorem~\ref{th:char:adm} it is clear (since any stable extension is admissible) that there is $\AF'$ with $(\AF,\emptyset)\leadsto^{\alpha_{\adm}}(\AF',S)$. As $S\cup S^{+}=\arguments$ it follows $\AF'=(\emptyset,\emptyset)$ and therefore $\beta_{\AF',S}=1$. Furthermore, for any $S$ with $(\AF,\emptyset)\leadsto^{\alpha_{\adm},\beta_{\st}}((\emptyset,\emptyset),S)$ it follows that $S$ is admissible and there is no $a\in \arguments$ with $a\notin S^{+}$ or $a\in S$. This is equivalent to stating that $S$ is stable.		
	\end{proof}
\end{theorem}
\begin{corollary}
	Let  $\AFcomplete$ be an abstract argumentation framework and $S\subseteq \arguments$. $S$ is stable if and only if either
	\begin{itemize}
		\item $S=\emptyset$ and $\arguments=\emptyset$ or
		\item $S=S_{1}\cup S_{2}$, $S_{1}\in\elc(\AF)$ and $S_{2}$ is stable in $\AF^{S_{1}}$.
	\end{itemize}
	\begin{proof}
		If $\arguments=\emptyset$ then $S=\emptyset$ is obviously the (only) stable extension. Assume $S$ is a non-empty stable extension of $\AF$. Due to Theorem~\ref{th:elc1} there are sets $S_1$, $S_2$ with $S=S_{1}\cup S_{2}$, $S_{1}\in\elc(\AF)$ and $S_{2}$ is admissible in $\AF^{S_{1}}$. Assume $S_2$ is not stable in $\AF^{S_{1}}=(\arguments',\attacks')$. Then there is $a\in \arguments'$ that is not attacked by $S_2$. Since $a\in\arguments'$, $a$ is also not attacked by $S_1$. It follows that $S$ is not a stable extension, in contradiction to the assumption. It follows that $S_2$ is stable in $\AF^{S_{1}}$.
		
		For the other direction, let $S=S_{1}\cup S_{2}$, $S_{1}\in\elc(\AF)$ and $S_{2}$ is stable in $\AF^{S_{1}}=(\arguments',\attacks')$. By Theorem~\ref{th:elc1}, $S$ is admissible. Assume $S$ is not stable, then there is $a\in \arguments$ that is not attacked by $S$. It follows that $a\in\arguments'$, so $a$ is also not attacked by $S_2$ in $\AF^{S_{1}}$, in contradiction to the assumption that $S_2$ is stable in $\AF^{S_{1}}$. It follows that $S$ is stable in $\AF$.
	\end{proof}
\end{corollary}
\begin{theorem}\label{th:pref}
	Preferred semantics is serialisable with $\alpha_{\adm}$ and
	\begin{align*}
		\beta_{\pr}(\AF,S)&=\left\{\begin{array}{ll}
							1 & \text{if $\elc(\AF)=\emptyset$}\\
							0 & \text{otherwise}
							\end{array}\right.
	\end{align*}
	\begin{proof}
		Let $S$ be preferred. Due to Theorem~\ref{th:char:adm} it follows that there is $\AF'$ with $(\AF,\emptyset)\leadsto^{\alpha_{\adm}}(\AF',S)$. If $\elc(\AF')\neq \emptyset$ then there is another admissible set $S'$ with $(\AF',S)\leadsto^{\alpha_{\adm}}(\AF'',S')$ and $S\subsetneq S'$, in contradiction to the assumption that $S$ is preferred. It follows $\elc(\AF')= \emptyset$ and therefore $(\AF,\emptyset)\leadsto^{\alpha_{\adm},\beta_{\pr}}(\AF',S)$. 
				
		For the other direction, let $(\AF,\emptyset)\leadsto^{\alpha_{\adm},\beta_{\pr}}(\AF',S)$. By Theorem~\ref{th:char:adm} it is clear that $S$ is admissible. Assume $S$ is not preferred, so there is admissible $S'$ with $S\subsetneq S'$. Define $S''=S'\setminus S$. We show now that $S''$ is an admissible set of $\AF'=\AF^{S}$. First, since $S'$ is conflict-free so is $S''$. Let now $a\in S''$. As $S'$ is admissible we have $a^{-}\subseteq (S')^{+}$. Let $a^{-}=X_{1}\cup X_{2}$ with disjoint sets $X_{1}$ and $X_{2}$ such that $X_{1}= a^{-}\cap S^+$ and $X_{2} = a^{-}\setminus X_{1}$. As $\AF'=\AF^{S}$, those attackers of $a$ in $X_{1}$ are not present in $\AF'$ anymore, so there is no need to defend against them. However, since $S'$ is admissible, $X_{2}\subseteq (S')^{+}$ and it follows $X_{2}\subseteq (S'')^{+}$ (as arguments in $X_{2}$ are not attacked by arguments in $S$). So $S''$ defends $a$ and $S''$ is therefore admissible.		
		As $S''$ is a (non-empty) admissible set of $\AF'$, it follows that $\elc(\AF')\neq\emptyset$. This contradicts the assumption that $(\AF,\emptyset)\leadsto^{\alpha_{\adm},\beta_{\pr}}(\AF',S)$ and it follows that $S$ is indeed a preferred extension.
	\end{proof}
\end{theorem}
\begin{corollary}
	Let  $\AFcomplete$ be an abstract argumentation framework and $S\subseteq \arguments$. $S$ is preferred if and only if either
	\begin{itemize}
		\item $S=\emptyset$ and $\elc(\AF)=\emptyset$ or
		\item $S=S_{1}\cup S_{2}$, $S_{1}\in\elc(\AF)$ and $S_{2}$ is preferred in $\AF^{S_{1}}$.
	\end{itemize}
	\begin{proof}
		If $\elc(\AF)=\emptyset$ then $S=\emptyset$ is obviously the only preferred extension, since it is the only admissible set. Assume $S$ is a non-empty preferred extension of $\AF$. Due to Theorem~\ref{th:elc1} there are sets $S_1$, $S_2$ with $S=S_{1}\cup S_{2}$, $S_{1}\in\elc(\AF)$ and $S_{2}$ is admissible in $\AF^{S_{1}}$. Assume $S_2$ is not preferred in $\AF^{S_{1}}$. Then there is admissible $S_2'$ with $S_2\subsetneq S_2'$. By Theorem~\ref{th:elc1}, $S_1\cup S_2'$ is admissible in $\AF$ and $S=S_1\cup S_2\subsetneq S_1\cup S_2'$, contradicting the fact that $S$ is preferred. It follows that $S_2$ is preferred in $\AF^{S_{1}}$.
		
		For the other direction, let $S=S_{1}\cup S_{2}$, $S_{1}\in\elc(\AF)$ and $S_{2}$ is preferred in $\AF^{S_{1}}$. By Theorem~\ref{th:elc1}, $S$ is admissible. Assume $S$ is not preferred, then there is admissible $S'$ with $S\subsetneq S'$. Since $S_1\subseteq S'$, $S'\setminus S_1$ must be completely contained in $\AF^{S_{1}}$ (otherwise $S'$ would not be conflict free) and $S_2\subsetneq S'\setminus S_1$. $S'\setminus S_1$ is also necessarily admissible in $\AF^{S_{1}}$, contradicting the fact that $S_2$ is preferred in $\AF^{S_{1}}$. It follows that $S$ is preferred.		
	\end{proof}
\end{corollary}
\begin{theorem}
	Strong admissibility semantics is serialisable with $\alpha_{\gr}$ and	$\beta_{\adm}$.
	\begin{proof}
		Let $S$ be a strongly admissible set. We show $(\AF,\emptyset)\leadsto^{\alpha_{\gr}}(\AF',S)$ (note that we do not have to consider $\beta_{\adm}$ as this function always returns $1$) by induction on the size of $S$.
		\begin{enumerate}
			\item $|S|=0$: trivial as $(\AF,\emptyset)\leadsto^{\alpha}(\AF,\emptyset)$ for any $\alpha$ via zero steps.
			\item $|S|=n$: Let $a\in S$ such that $S'=S\setminus\{a\}$ is strongly admissible (the existence of such $a$ is guaranteed as a direct corollary of Theorem~5 in \cite{DBLP:journals/argcom/CaminadaD19} and the definition of strong admissibility). By induction hypothesis $(\AF,\emptyset)\leadsto^{\alpha_{\gr}}(\AF',S')$. As $S$ is strongly admissible, it follows that $a$ cannot be attacked in $\AF'$ (otherwise $a$ is not defended by (a subset of) S'). So we have $\{a\}\in\uelc(\AF')$ and $(\AF,\emptyset)\leadsto^{\alpha_{\gr}}(\AF',S')\leadsto^{\alpha_{\gr}}(\AF'',S)$.
		\end{enumerate}
		For the other direction, let $(\AF,\emptyset)\leadsto^{\alpha_{\gr}}(\AF',S)$. We show that $S$ is strongly admissible by induction on the size of $S$.
		\begin{enumerate}
			\item $|S|=0$: the empty set is by definition strongly admissible.
			\item $|S|=n$: Consider the final step in the construction of $S$, i.\,e., $(\AF,\emptyset)\leadsto^{\alpha_{\gr}}(\AF',S')\leadsto^{\alpha_{\gr}}(\AF'',S)$. As $\alpha_{gr}$ returns only singleton sets, we have $S=S'\cup\{a\}$ for some $a\in \arguments$. As $a$ is unattacked in $\AF''$, $S'$ defends $a$ in $\AF$. By induction hypothesis, $S'$ is strongly admissible, showing that $S$ is strongly admissible.\qedhere
		\end{enumerate}
	\end{proof}
\end{theorem}
\begin{corollary}
	Let  $\AFcomplete$ be an abstract argumentation framework and $S\subseteq \arguments$. $S$ is strongly admissible if and only if either
	\begin{itemize}
		\item $S=\emptyset$ or
		\item $S=S_{1}\cup S_{2}$, $S_{1}\in\uelc(\AF)$ and $S_{2}$ is strongly admissible in $\AF^{S_{1}}$.
	\end{itemize}
	\begin{proof}
		Since $\emptyset$ is always strongly admissible, we only consider the second case. 
		Assume $S$ is a non-empty strongly admissible set. Since $S\subseteq E_{gr}$, where $E_{gr}$ is the grounded extension of $\AF$, there is an unattacked $a\in S$ and $\{a\}\in \uelc(\AF)$. Let $S_2 = S\setminus\{a\}$ and $b\in S_2$. Since $S$ is strongly admissible in $\AF$, there is $S''\subseteq S\setminus\{b\}$ that is strongly admissible and defends $b$. If $a\in S''$ then $S''\setminus\{a\}$ is strongly admissible in $\AF^{\{a\}}$ and defends $b$. If $a\notin S''$ then $S''$ remains strongly admissible in $\AF^{\{a\}}$ and defends $b$. In any case, it follows that $S_2$ is strongly admissible in $\AF^{\{a\}}$.
	
		The other direction is analogous.
	\end{proof}
\end{corollary}
\setcounter{theorem}{9}
\begin{theorem}
	For every $E$ with $(\AF,\emptyset)\leadsto^{\alpha_{0},\beta_{0}}(\AF',E)$
	\begin{enumerate}
		\item $E\subseteq E'$ for some preferred extension $E'$ and
		\item $E_{\id}\subseteq E$ for the ideal extension $E_{\id}$.
	\end{enumerate}
	\begin{proof}~
		\begin{enumerate}
			\item Note that $\alpha_{0}(X,Y,Z)\subseteq \alpha_{\adm}(X,Y,Z)$ for all $X,Y,Z$. So if $(\AF,\emptyset)\leadsto^{\alpha_{0}}(\AF',E)$ then $(\AF,\emptyset)\leadsto^{\alpha_{\adm}}(\AF',E)$. Furthermore, $(\AF',E)\leadsto^{\alpha_{\adm},\beta_{\pr}}(\AF'',E')$ eventually with a preferred extension $E'$ due to Theorem~\ref{th:pref}. This shows $E\subseteq E'$.
			\item	Let $E_{\id}$ be the ideal extension of $\AF$ and let $E_{\id}=S_{1}\cup\ldots\cup S_{n}$ with $S_{i}$ being an initial set of $\AF^{S_{1}\cup\ldots\cup S_{i-1}}$ for all $i=1,\ldots,n$ (this representation exists due to Corollary~\ref{cor:rec} and the fact that $E_{\id}$ is admissible). 
			
			Let $E$ with $(\AF,\emptyset)\leadsto^{\alpha_{0},\beta_{0}}(\AF^{E},E)$. Assume that $E_{\id}\not\subseteq E$ and let $k\in\{1,\ldots,n\}$ be the smallest integer  such that $S_{k}\not\subseteq E$. Let $\hat{S}_{k}=E\setminus S_{k}$. We show now that $\hat{S}_{k}$ is admissible in $\AF^{E}=(\arguments',\attacks')$:
			\begin{enumerate}
				\item $\hat{S}_{k}\subseteq \arguments'$: For the sake of contradiction, assume there is $a\in\hat{S}_{k}$ with $a\notin \arguments'$. Due to $\hat{S}_{k}=E\setminus S_{k}$ it follows that $a$ is attacked by $E$. Then the admissible set $E$ attacks $E_{\id}$, contradicting the fact that $E_{\id}$ is the ideal extension.
				\item $\hat{S}_{k}$ is conflict-free: clear since $\hat{S}_{k}\subseteq E_{\id}$.
				\item $\hat{S}_{k}$ defends all its elements (in $\AF^{E}$): recall that $S_{1}\cup\ldots\cup S_{k}$ is an admissible set in $\AF$ (due to Corollary~\ref{cor:rec}) and that $S_{1}\cup\ldots\cup S_{k-1}\cup (S_{k}\setminus \hat{S}_{k})\subseteq E$. Let $a$ be an attacker of $\hat{S}_{k}$ in $\AF$ and $b\in S_{1}\cup\ldots\cup S_{k}$ that attacks $a$. Then either $b\in E$ (meaning that $a\notin \arguments'$ and $\hat{S}_{k}$ does not need to defend against $a$ in $\AF^{E}$) or $b\in \hat{S}_{k}$ (meaning that $\hat{S}_{k}$ defends against $a$ in $\AF^{E}$). 
			\end{enumerate}
			 It follows that $\hat{S}_{k}$ is admissible in $\AF^{E}$. Then there must be an initial set $\hat{S}'_{k}\subseteq\hat{S}_{k}$. Assume $\hat{S}'_{k}$ is challenged by another initial set $T$. Then $T\cup E$ would be an admissible set that attacks $E_{\id}$. It follows $\hat{S}'_{k}$ is unattacked or unchallenged in $\AF^{E}$. This contradicts the fact that $(\AF,\emptyset)\leadsto^{\alpha_{0},\beta_{0}}(\AF^{E},E)$. Therefore we have 
			 	$E_{\id}\subseteq E$.\qedhere
		\end{enumerate}
	\end{proof}
\end{theorem}
\begin{lemma}\label{lem:subadm}
	Let $S$ be conflict-free and $a\in S$. Deciding whether there is an admissible set $S'\subseteq S$ with $a\in S'$ can be decided in polynomial time.
	\begin{proof}
		In polynomial time we can check first whether $S$ is already admissible. If not, define $S_{1}$ via
		\begin{align*}
			S_{1} & = F_{\AF}(S)\cap S
		\end{align*}
		Note that $S_{1}\subsetneq S$ (if $S_{1}=S$ then $S$ would already have been admissible). Furthermore, all $a\in S\setminus S_{1}$ are not defended by arguments in $S$ and can therefore not be a member of any admissible set $S'\subseteq S$. It follows that, if there is an admissible set $S'\subseteq S$ with $a\in S'$ then $S'\subseteq S_{1}$. So if $a\notin S_{1}$ or ($a\in S_{1}$ and $S_{1}$ is admissible), we are finished. Otherwise define $S_{2}$ via 
		\begin{align*}
			S_{2} & = F_{\AF}(S_{1})\cap S_{1}
		\end{align*}
		and continue as before. Note that moving from $S_{i}$ to $S_{i+1}$ at least one argument is discarded (otherwise we have found an admissible set). So we have to compute at maximum $S_{1},\ldots, S_{|S|}$ and all computations are polynomial.
	\end{proof}
\end{lemma}
\begin{proposition}\label{prop:elc:poly}
	$\textit{Ver}_{\elc}$ is in \textsf{P}.
	\begin{proof}
		In polynomial time we can check first whether the input $S$ is admissible. Then, for each $a,b\in S$ with $a\neq b$ we can test whether $S\setminus\{b\}$ contains an admissible set including $a$ (see Lemma~\ref{lem:subadm}). If this is the case for one pair $a,b$, $S$ cannot be an initial set. If this is not the case for any $a,b$ then $S$ is an initial set. All checks are in polynomial time.	
	\end{proof}
\end{proposition}
\begin{proposition}
	$\textit{Exists}_{\elc}$ is \textsf{NP}-complete.
	\begin{proof}
		Equivalence of $\textit{Exists}_{\elc}$ and $\textit{Exists}^{\neg \emptyset}_{\adm}$ follows from the fact that every non-empty admissible set contains an initial set \cite{Xu:2016}. $\textit{Exists}^{\neg \emptyset}_{\adm}$ is \textsf{NP}-complete \cite{Dvorak:2018}.
	\end{proof}
\end{proposition}
\begin{proposition}\label{prop:unique:elc}
	$\textit{Unique}_{\elc}$ is \textsf{DP}-complete.
	\begin{proof}
		Let $\AF$ be the input argumentation framework. Note that $\textit{Unique}_{\elc}$ can be solved by solving the two problems: 
			\begin{enumerate}
				\item decide whether $\AF$ has at least one initial set and
				\item decide whether $\AF$ has at most one initial set.
			\end{enumerate}
		Problem 1 is $\textit{Exists}_{\elc}$ and therefore \textsf{NP}-complete. The complement of problem 2 can be solved by non-deterministically guessing two different sets $S_{1}$ and $S_{2}$ and verifying that both are initial sets. Problem 2 is therefore in \textsf{coNP} and this shows \textsf{DP}-membership of $\textit{Unique}_{\elc}$.
		
		For hardness, we provide a reduction from the problem $\textit{Unique}_{\st}$, i.\,e., the problem of deciding whether an argumentation framework has a unique stable extension, cf.\ \cite{Dvorak:2018,Dimopoulos:1996}.
		For that, we directly use construction $Tr_4$ from \cite{Dvorak:2011}, which translates (with polynomial overhead) an argumentation framework $\AF$ into an argumentation framework $Tr_4(\AF)$ such that $\st(\AF)=\adm(Tr_4(\AF))\setminus\{\emptyset\}$, cf. Lemma~5 and Theorem~4 in \cite{Dvorak:2011}. Since for every pair of stable extensions $E_1,E_2$ it holds $E_1\not\subseteq E_2$ and $E_2\not\subseteq E_1$, it follows that $E_1,E_2\in\adm(Tr_4(\AF))\setminus\{\emptyset\}$, $E_1\not\subseteq E_2$ and $E_2\not\subseteq E_1$, and therefore $\adm(Tr_4(\AF))\setminus\{\emptyset\}=\elc(Tr_4(\AF))$. It follows $\st(\AF)=\elc(Tr_4(\AF))$ and $|\st(\AF)|=1$ if and only if $|\elc(Tr_4(\AF))|=1$ and therefore the claim.			
	\end{proof}
\end{proposition}
\begin{proposition}\label{prop:cred:elc}
	$\textit{Cred}_{\elc}$ is \textsf{NP}-complete.
	\begin{proof}
			For \textsf{NP}-membership consider the following algorithm. Upon input $a\in\arguments$ we guess a set $S\subseteq\arguments$ with $a\in S$ and verify in polynomial time that $S$ is an initial set (see Proposition~\ref{prop:elc:poly}). It follows that $a$ is credulously accepted wrt.\ initial sets. This shows \textsf{NP}-membership.
		
		For \textsf{NP}-hardness we do a reduction from $\textit{Cred}_{\st}$, i.\,e., the problem of deciding whether an argument is credulously accepted wrt.\ stable semantics. We use the same reduction as in the proof of Proposition~\ref{prop:unique:elc}, namely the construction $Tr_4$ from \cite{Dvorak:2011}. We already established in the proof of Proposition~\ref{prop:unique:elc} that $\st(\AF)=\elc(Tr_4(\AF))$. It follows that an argument $a$ is credulously accepted wrt.\ stable semantics in $\AF$ if and only if it is credulously accepted wrt.\ initial sets in $Tr_4(\AF)$.
	\end{proof}
\end{proposition}
\begin{proposition}
	$\textit{Skept}_{\elc}$ is \textsf{coNP}-complete.
	\begin{proof}
		For \textsf{coNP}-membership consider the following algorithm, which solves the complement problem in \textsf{NP}. Upon input $a\in\arguments$ we guess a set $S\subseteq\arguments$ with $a\notin S$ and verify in polynomial time that $S$ is an initial set (see Proposition~\ref{prop:elc:poly}). It follows that $a$ is not skeptically  accepted wrt.\ initial sets. This shows \textsf{coNP}-membership for $\textit{Skept}_{\elc}$.
		
		For \textsf{coNP}-hardness we do a reduction from $\textit{Skept}_{\st}$, i.\,e., the problem of deciding whether an argument is skeptically accepted wrt.\ stable semantics. We use the same reduction as in the proof of Proposition~\ref{prop:unique:elc}, namely the construction $Tr_4$ from \cite{Dvorak:2011}. We already established in the proof of Proposition~\ref{prop:unique:elc} that $\st(\AF)=\elc(Tr_4(\AF))$. It follows that an argument $a$ is skeptically accepted wrt.\ stable semantics in $\AF$ if and only if it is skeptically accepted wrt.\ initial sets in $Tr_4(\AF)$.
	\end{proof}
\end{proposition}

\begin{proposition}\label{prop:ver:uelc}
	$\textit{Ver}_{\uelc}$, $\textit{Exists}_{\uelc}$, $\textit{Unique}_{\uelc}$, $\textit{Cred}_{\uelc}$, and $\textit{Skept}_{\uelc}$ are in \textsf{P}.	
	\begin{proof}
		Note that all these problems only have to make a simple check on the input:
		\begin{itemize}
			\item $\textit{Ver}_{\uelc}$: Verifying whether a single argument is unattacked is in \textsf{P}.
			\item $\textit{Exists}_{\uelc}$: Checking whether there is an unattacked argument in a given input framework $\AF$ is in \textsf{P}.
			\item $\textit{Unique}_{\uelc}$: Checking whether there is a single unattacked argument in a given input framework $\AF$ is in \textsf{P}.
			\item $\textit{Cred}_{\uelc}$: this is equivalent to $\textit{Ver}_{\uelc}$  with input $\AF$ and $\{a\}$.
			\item $\textit{Skept}_{\uelc}$: checking whether $a$ is the only unattacked argument in $\AF$ is in \textsf{P}.\qedhere
		\end{itemize}
	\end{proof}
\end{proposition}

\begin{proposition}\label{prop:ver:ucelc}
	$\textit{Ver}_{\ucelc}$ is \textsf{coNP}-complete.	
	\begin{proof}
		For \textsf{coNP}-membership, we consider the complement problem of verifying that input $S$ is \emph{not} an unchallenged initial set in \textsf{NP}. We first check in polynomial time whether the input $S$ is an initial set at all, see Proposition~\ref{prop:elc:poly}. Then we check whether $S$ is an unattacked initial set, see Proposition~\ref{prop:ver:uelc}. If it is not, we guess another set $S'$ with $S'\attacks S$ and verify in polynomial time that $S'$ is an initial set. This shows that $S$ is not an unchallenged initial set in \textsf{NP}.
		
		For \textsf{coNP}-hardness, we do a reduction from 3UNSAT,  i.\,e., the problem of deciding whether a propositional formula in conjunctive normal form with exactly three literals per clause is unsatisfiable. For that, we extend Reduction 3.6 from \cite{Dvorak:2018}. Let $\phi$ be an instance of 3UNSAT in set notation, i.\,e., $\phi=\{C_{1}, \ldots, C_{n}\}$ and $C_{i}=\{l_{1,i},l_{2,i},l_{3,i}\}$ with literals $l_{1,i},l_{2,i},l_{3,i}$ from a set of atoms $At$, for $i=1,\ldots,n$. Define an abstract argumentation framework $\AF'_{\phi}=(\arguments'_{\phi},\attacks'_{\phi})$ via
		\begin{align*}
			\arguments'_{\phi} &= \{\phi,\tilde{\phi},\psi\} \cup \{C_{1},\ldots,C_{n}\}\cup \{a,\neg a\mid a\in At\}\\
			\attacks'_{\phi} & = \{(C_{1},\phi),\ldots,(C_{n},\phi)\}\cup\\
						&\hspace*{0.47cm}\{(l,C_{i}) \mid l\in C,i\in\{1,\ldots,n\}\}\cup\\
						&\hspace*{0.47cm}\{(a,\neg a),(\neg a, a) \mid a\in At\}\cup\\
						&\hspace*{0.47cm}\{(\tilde{\phi},a),(\tilde{\phi},\neg a)\mid a\in At\}\cup\\
						&\hspace*{0.47cm}\{(\phi,\tilde{\phi}),(\phi,\psi),(\psi,\phi)\}
		\end{align*}
		Figure~\ref{fig:exreduct2} shows an example of the reduction. We first show that there is an initial set containing $\phi$ if and only if $\phi$ is satisfiable.
		For that, assume first that $\phi$ is satisfiable and let $I:At\rightarrow \{\textsf{true},\textsf{false}\}$ be a model of $\phi$. Consider the set 
		\begin{align*}
			S_{I} & = \{ a \mid I(a)=\textsf{true}\}\cup\{ \neg a \mid I(a)=\textsf{false}\}\cup \{\phi\}
		\end{align*}
		First observe that $S_{I}$ is conflict-free: as $I$ is an interpretation it is not the case that $a,\neg a\in S$ for some $a\in At$. Furthermore, there are no attacks between any argument from $\{a,\neg a\mid a\in At\}$ to $\phi$ and vice versa. Now observe that $S_{I}$ is admissible (in fact $S_{I}$ is stable):
		\begin{enumerate}
			\item each $a\in S$ defends itself against $\neg a$ (for $a\in At$),
			\item each $\neg a\in S$ defends itself against $a$ (for $a\in At$),
			\item each $C_{i}$ is attacked by some $a\in S$ or $\neg a \in S$ (since $I$ is a model, every clause is satisfied), and
			\item $\tilde{\phi}$ is attacked by $\phi$.			
		\end{enumerate}
		However, $S_{I}$ is not necessarily an initial set. Consider the example in Figure~\ref{fig:exreduct2} again: here, $I$ with $I(a)=I(b)=I(c)=\textsf{false}$ is a model of $\phi$ and we have 
		\begin{align*}
			S_{I} & = \{\neg a, \neg b, \neg c, \phi\}
		\end{align*}
		Furthermore, $S_{I}'=\{\neg a, \neg b, \phi\}\subseteq S_{I}$ is also admissible (this happens when the truth value of one or more atoms does not matter for satisfiability). However, due to the fact that every non-empty admissible set contains an initial set, there is always an initial set $S_{I}'\subseteq S_{I}$ and $S_{I}'$ must always contain $\phi$ as this is the only argument defending all arguments in $\{a,\neg a\mid a\in At\}$. It follows that if $\phi$ is satisfiable then there is an initial set containing $\phi$. The other direction is analogous.		
		
		We now claim that $\phi$ is unsatisfiable if and only if $\{\psi\}$ is an unchallenged initial set. First, it is clear that $\{\psi\}$ is an initial set since $\psi$ counterattacks the only attack. Moreover, we established above that $\phi$ is satisfiable if and only if there is an initial set $S$ containing $\phi$. So if $\phi$ is satisfiable $\{\psi\}$ is challenged by $S$. If $\phi$ is unsatisfiable then $\{\psi\}$ is clearly unchallenged.
\begin{figure}[t]
\begin{center}
\begin{tikzpicture}[node distance=0.35cm]

	\node[args](x){$\phi$};
	
	\node[args, below=of x, yshift=-0.4cm, xshift=-3cm](c1){$c_{1}$};
	\node[args, below=of x,yshift=-0.4cm](c2){$c_{2}$};
	\node[args, below=of x,yshift=-0.4cm, xshift=3cm](c3){$c_{3}$};
	
	\node[args, below=of c1,yshift=-0.4cm, xshift=-0.6cm](a){$a$};
	\node[args, below=of c1,yshift=-0.4cm, xshift=0.6cm](an){$\neg a$};

	\node[args, below=of c2, yshift=-0.4cm,xshift=-0.6cm](b){$b$};
	\node[args, below=of c2, yshift=-0.4cm,xshift=0.6cm](bn){$\neg b$};
	
	\node[args, below=of c3, yshift=-0.4cm,xshift=-0.6cm](c){$c$};
	\node[args, below=of c3, yshift=-0.4cm,xshift=0.6cm](cn){$\neg c$};
	
	\node[args, below=of b, xshift=0.6cm, yshift=-0.4cm](y){$\tilde{\phi}$};
	
	\node[args, above=of c1, yshift=0.4cm](psi){$\psi$};
	
	\path(c1) edge [->] (x);
	\path(c2) edge [->] (x);
	\path(c3) edge [->] (x);
	
	\path(a) edge [->, bend left] (an);
	\path(an) edge [->, bend left] (a);
	\path(b) edge [->, bend left] (bn);
	\path(bn) edge [->, bend left] (b);
	\path(c) edge [->, bend left] (cn);
	\path(cn) edge [->, bend left] (c);

	\path(a) edge [->] (c1);
	\path(bn) edge [->] (c1);
	\path(c) edge [->] (c1);
	\path(an) edge [->] (c2);
	\path(bn) edge [->] (c2);
	\path(c) edge [->] (c2);
	\path(an) edge [->] (c3);
	\path(b) edge [->] (c3);
	\path(cn) edge [->] (c3);
	
	\path(y) edge [->] (a);
	\path(y) edge [->] (an);
	\path(y) edge [->] (b);
	\path(y) edge [->] (bn);
	\path(y) edge [->] (c);
	\path(y) edge [->] (cn);
	
	\path(psi) edge [->, bend left=15] (x);
	\path(x) edge [->, bend left=15] (psi);
	
	\node[right=of x, xshift=3.6cm, minimum size=0cm, inner sep=0cm](x1){};
	\node[right=of y, xshift=3.6cm, minimum size=0cm, inner sep=0cm](x2){};
	\path(x) edge [-] (x1);
	\path(x1) edge [-] (x2);
	\path(x2) edge [->] (y);

\end{tikzpicture}
\end{center}
\caption{The argumentation framework $\AF'_{\phi}$ for $\phi=\{\{a, \neg b, c\},\{\neg a , \neg b, c\},\{\neg a, b, \neg c\}\}$.}
\label{fig:exreduct2}
\end{figure}
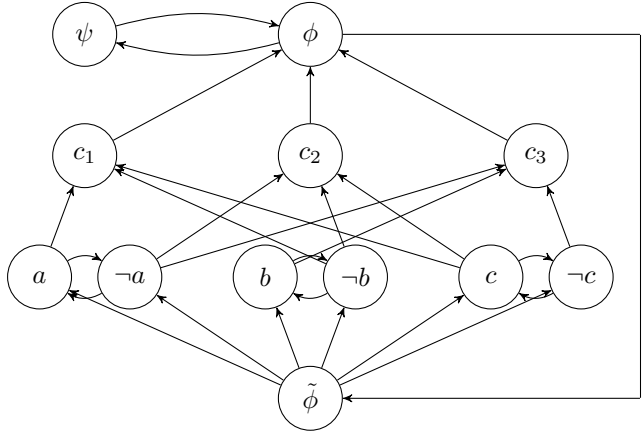	
	\end{proof}
\end{proposition}

\begin{proposition}\label{prop:exists:ucelc}
	$\textit{Exists}_{\ucelc}$ is in $\textsf{P}^{\textsf{NP}}_{\parallel}$-complete.	
	\begin{proof}
	In order to show $\textsf{P}^{\textsf{NP}}_{\parallel}$-completeness, we use a characterisation of $\textsf{P}^{\textsf{NP}}_{\parallel}$ from \cite{Chang:1995}.\footnote{I am very grateful to an anonymous reviewer for pointing out that characterisation.} More precisely, Theorem~9 of \cite{Chang:1995} establishes that a problem $X$ is $\textsf{P}^{\textsf{NP}}_{\parallel}$-complete if and only if
		\begin{enumerate}
			\item $X\in \textsf{P}^{\textsf{NP}}_{\parallel}$,
			\item $X$ is $\textsf{NP}$-hard,
			\item $X$ is $\textsf{coNP}$-hard,
			\item Two problem instances $i_1$ and $i_2$ of $X$ can be polynomially reduced to a problem instance $i_3$ of $X$ such that $i_3$ is a positive instance if and only if both $i_1$ and $i_2$ are positive instances.
			\item A set of problem instances $I=\{i_1,\ldots,i_k\}$ of $X$ can be polynomially reduced to a problem instance $i$ of $X$ such that $i$ is a positive instance if and only if there is at least one positive instance in $I$.
		\end{enumerate}
		We now show that properties 1--5 above hold for the problem $\textit{Exists}_{\ucelc}$.
	\begin{enumerate}
		\item 	
		For $\textsf{P}^{\textsf{NP}}_{\parallel}$-membership, consider the following algorithm\footnote{Note that this algorithm is inspired by an algorithm for determining the ideal extension, cf. \cite{Dung:2007,Dunne:2009}}. Let $\AFcomplete$ be the input argumentation framework.
		\begin{enumerate}
			\item[1.] For each argument $a\in\arguments$, check whether $a$ is not attacked by an initial set
			\item[2.] For each argument $a\in\arguments$, check whether $a$ is contained in an initial set
			\item[3.] Let $M$ be the set of arguments for which both checks 1 and 2 were positive
			\item[4.] Remove all unattacked arguments from $M$, yielding a new set $M'$
			\item[5.] Compute the maximal admissible set $M''$ in $M'$ (which is uniquely determined)
			\item[6.] If $M''\neq\emptyset$ return \textsc{Yes}, otherwise return \textsc{No}
		\end{enumerate}
		First observe that the above algorithm runs in $\textsf{P}^{\textsf{NP}}_{\parallel}$. All checks in steps 1 and 2 can be solved by an \textsf{NP}-oracle:
		\begin{enumerate}
			\item The check in step 1 is a decision problem in \textsf{coNP} as the complement problem (i.\,e., checking whether $a$ is attacked by an initial set) can be solved by guessing a set $S$ that attacks $a$ and verifying in polynomial time (see Proposition~\ref{prop:elc:poly}) that it is an initial set.
			\item The check in step 2 is a decision problem in \textsf{NP}: it can be solved by guessing a set $S$ that contains $a$ and verifying in polynomial time (see Proposition~\ref{prop:elc:poly}) that it is an initial set.
		\end{enumerate}
		All checks in step 1 and 2 are non-adaptive, so they can be done in parallel (and there are linearly many of them). For step 4, observe that identifying unattacked arguments can be done in polynomial time. For step 5, note that $M'$ is conflict-free (if there would be two arguments $a,b\in M'$ with $a\attacks b$, it means that both $a$ and $b$ are in some (possibly different) initial sets and that $b$ is in some initial set that is attacked by some other initial set, which cannot be due to step 1). Determining the maximal admissible set in $M'$ can be done similarly as in the proof of Lemma~\ref{lem:subadm} in polynomial time.
		
		Now we claim that the above algorithm returns \textsc{Yes} if and only if the answer to $\textit{Exists}_{\ucelc}$ upon input $\AF$ is \textsc{Yes}.
		\begin{itemize}
			\item ``$\Rightarrow$'': Let $M''\neq \emptyset$ be the admissible set computed in step 5 and let $S\subseteq M''$ be an initial set contained in $M''$ (which necessarily exists since $M''$ is admissible and non-empty). Observe that $S$ is not an unattacked initial set as we removed all unattacked arguments in step 4. Assume $S$ is a challenged initial set. Then there exists another initial set $S'$ which attacks some argument $b\in S$. This is in contradiction to the fact that $S\subseteq M'\subseteq M$ and $M$ contains only arguments that are not attacked by an initial set (see step 1). So $S$ is an unchallenged initial set and the answer to $\textit{Exists}_{\ucelc}$ upon input $\AF$ is \textsc{Yes}.
			\item ``$\Leftarrow$'': Assume the answer to $\textit{Exists}_{\ucelc}$ upon input $\AF$ is \textsc{Yes}. Then there exists an unchallenged initial set $S$. By definition, every argument $a\in S$ is contained in an initial set and not attacked by an initial set, so $S\subseteq M$ in step 3 of the above algorithm. Since $S$ is unchallenged (but not unattacked), every argument in $S$ is attacked and we have $S\subseteq M'$ in step 4. As $S$ is admissible (and in no conflict with any other argument in $M'$) we also have $S\subseteq M''$ in step 5 of the above algorithm. As $S\neq\emptyset$ it follows $M''\neq \emptyset$ as well and the algorithm return \textsc{Yes}.
		\end{itemize}
		In conclusion, $\textit{Exists}_{\ucelc}$ is in $\textsf{P}^{\textsf{NP}}_{\parallel}$.		
		\item We show \textsf{DP}-hardness (which entails \textsf{NP}-hardness) instead.
			For that we use the same reduction from $\textit{Unique}_{\st}$ as in the proof of Proposition~\ref{prop:unique:elc}, i.\,e., the construction $Tr_4$ from \cite{Dvorak:2011}. In particular, observe that if $Tr_4(\AF)$ has exactly one initial set $S$ then $S$ is unchallenged. Furthermore, if $Tr_4(\AF)$ has no initial sets then it obviously also has no unchallenged initial sets. If $Tr_4(\AF)$ has at least two initial sets, then all these initial sets are challenged (as they are all stable extensions of the original framework and two different stable extensions necessarily attack each other). This shows \textsf{DP}-hardness of $\textit{Exists}_{\ucelc}$.	
		\item Since $\textit{Exists}_{\ucelc}$ is \textsf{DP}-hard (see above) it is also \textsf{coNP}-hard.
		\item Let $\AF_1=(\arguments_1,\attacks_1)$ and $AF_2=(\arguments_2,\attacks_2)$ be two argumentation frameworks and assume $\arguments_1\cap \arguments_2=\emptyset$ (otherwise rename arguments accordingly). Construct $\AF_3=(\arguments_3,\attacks_3)$ as follows (let $a_1,a_2$ be fresh arguments):
			\begin{align*}
				\arguments_3 & = \arguments_1\cup\arguments_2\cup\{a_1,a_2\}\\
				\attacks_3 & = \attacks_1\cup\attacks_2\cup\{(a_1,a_1),(a_2,a_2)\}\\
							&\quad \cup\{(a,a_1)\mid a\in \arguments_1, a^-\neq \emptyset\}
							 \cup\{(a_1,b)\mid b\in \arguments_2, b^-\neq \emptyset\}\\
							&\quad \cup\{(b,a_2)\mid b\in \arguments_2, b^-\neq \emptyset\}
							 \cup\{(a_2,a)\mid a\in \arguments_1, b^-\neq \emptyset\}
			\end{align*}
			The intuition behind the above construction is that the two frameworks $\AF_1$ and $AF_2$ are arranged in a circle where every (already attacked) argument of $\AF_1$ attacks $a_1$, $a_1$ attacks every (already attacked) argument of $\AF_2$, which in turn all attack $a_2$, which in turn attacks all (already attacked) arguments of $\AF_1$. Obviously, the construction of $\AF_3$ is polynomial in the size of $\AF_1$ and $\AF_2$.
			
			We now show that $\AF_3$ has an unchallenged initial set if and only if both $\AF_1$ and $\AF_2$ have unchallenged initial sets. Let $M$ be an unchallenged initial set of $\AF_3$. Since $a_1$ and $a_2$ attack themselves, $a_1,a_2\notin M$. Assume $M\subseteq \arguments_1$. Since $M$ is unchallenged and not unattacked, all $a\in M$ are attacked. It follows that $a_2$ attacks $M$. Since only arguments in $\AF_2$ attack $a_2$, $M$ cannot defend itself from $a_2$. It follows that $\arguments_2\cap M\neq \emptyset$. For the same reason and using $a_1$ instead of $a_2$, it follows $\arguments_1\cap M\neq \emptyset$. Let $M_1=M\cap\arguments_1$ and $M_2=M\cap\arguments_2$. Assume $M_1$ is challenged in $\AF_1$ and let $M_1'$ be an initial set in conflict with $M_1$ in $\AF_1$. Then $M_1'\cup M_2$ is also an initial set of $\AF_3$ and in conflict with $M$, contradicting the assumption that $M$ is unchallenged. It follows that both $M_1$ and (with the same argument) $M_2$ are unchallenged in $AF_1$ and $\AF_2$, respectively. For the other direction, given that $M_1$ and $M_2$ are unchallenged initial sets of $\AF_1$ and $\AF_2$, respectively, it is also clear that $M_1\cup M_2$ is an unchallenged initial set of $\AF_3$.
		\item Let $\AF_1=(\arguments_1,\attacks_1),\ldots, \AF_n=(\arguments_n,\attacks_n)$ be argumentation frameworks and assume $\arguments_i\cap \arguments_j=\emptyset$ for all $i,j=1,\ldots, n$ and $i\neq j$ (otherwise rename arguments accordingly). Observe that $AF=(\arguments_1\cup\ldots\cup\arguments_n,\attacks_1\cup\ldots\cup\attacks_n)$ has an unchallenged initial set $M$ if and only if at least one of $\AF_1,\ldots,\AF_n$ has an unchallenged initial set (since necessarily $M\subseteq \arguments_i$ for some $i$ due to the disconnectedness of $\AF$).
  \qedhere
	\end{enumerate}		
	\end{proof}
\end{proposition}

\begin{proposition}
	$\textit{Unique}_{\ucelc}$ is in $\textsf{P}^{\textsf{NP}}_{\parallel}$ and \textsf{DP}-hard.	
	\begin{proof}
	For $\textsf{P}^{\textsf{NP}}_{\parallel}$-membership, we use a similar algorithm as in the proof of Proposition~\ref{prop:exists:ucelc}. Let $\AFcomplete$ be the input argumentation framework.
		\begin{enumerate}
			\item[1.] For each argument $a\in\arguments$, check whether $a$ is not attacked by an initial set
			\item[2.] For each argument $a\in\arguments$, check whether $a$ is contained in an initial set
			\item[3.] Let $M$ be the set of arguments for which both checks 1 and 2 were positive
			\item[4.] Remove all unattacked arguments from $M$, yielding a new set $M'$
			\item[5.] Compute the maximal admissible set $M''$ in $M'$ (which is uniquely determined)
			\item[6.] If $M''=\emptyset$ return \textsc{No}
			\item[7.] For each argument $a\in M''$, let $M_{a}=M''\setminus \{a\}$
			\item[8.] For each argument $a\in M''$, let $M'_{a}$ be the maximal admissible set in $M_{a}$ (which is uniquely determined)
			\item[9.] If for all $a\in M''$, $M'_{a}\neq \emptyset$, return \textsc{No}
			\item[10.] Let $S=\{a\mid M'_{a}= \emptyset\}$
			\item[11.] If $S$ is an initial set return \textsc{Yes}, otherwise return \textsc{No}
		\end{enumerate}
		First observe that the above algorithm runs in $\textsf{P}^{\textsf{NP}}_{\parallel}$. Steps 1--6 run in $\textsf{P}^{\textsf{NP}}_{\parallel}$ as already shown in the proof of Proposition~\ref{prop:exists:ucelc}. Furthermore, steps 7--11 run in (deterministic) polynomial time (in particular, step 8 runs in polynomial time by leveraging a similar algorithm as in the proof of Lemma~\ref{lem:subadm} and step 11 because of Proposition~\ref{prop:elc:poly}). 
		
		We now claim that the above algorithm returns \textsc{Yes} if and only if $\AF$ has a unique unchallenged initial set.
		\begin{itemize}
			\item ``$\Rightarrow$'': If the algorithm returns \textsc{Yes}, we have obviously found an initial set $S$ in step 11. As $S\subseteq M''$ we also have that $S$ is an unchallenged initial set (see the proof of Proposition~\ref{prop:exists:ucelc}). Assume there exists an unchallenged initial set $S'$ with $S'\neq S$. Note that both $S\not\subseteq S'$ and $S'\not\subseteq S$ since both are initial sets and $S'\subseteq M''$ (again, see the proof of Proposition~\ref{prop:exists:ucelc}). Let $x\in S\setminus S'$. Then $M'_{x}\neq \emptyset$ as $M_{x}$ completely contains $S'$. This contradicts the fact that $x\in S$ due to step 10.
				It follows that $\AF$ has the unique unchallenged initial set $S$.
			\item ``$\Leftarrow$'': Assume $\AF$ has the unique unchallenged initial set $M$. By the argumentation in the proof of Proposition~\ref{prop:exists:ucelc} we have that $M\subseteq M''$ in step 6. Then for all $a\in M''$ we have that 
			\begin{itemize}
				\item $M'_{a}=\emptyset$ if and only if $a\in M$ since $M_{a}\neq \emptyset$ would imply that there is another unchallenged initial set contained in $M_{a}$.
				\item $M'_{a}\neq\emptyset$ if and only if $a\notin M$ as $M$ is contained in $M_{a}$.
			\end{itemize}
			By the definition of $S$ in step 10 it follows $S=M$. As $M$ is an initial set, the algorithm returns \textsc{Yes} in step 11.
		\end{itemize}	
	
		For showing \textsf{DP}-hardness we use the same reduction from $\textit{Unique}_{\st}$ as in the proof of Proposition~\ref{prop:unique:elc}, i.\,e., the construction $Tr_4$ from \cite{Dvorak:2011}. In particular, note that $\AF$ has exactly one stable extension if and only if $Tr_4(\AF)$ has exactly one (unchallenged) initial set.
	\end{proof}
\end{proposition}

\begin{proposition}\label{prop:ucelc:cred}
	$\textit{Cred}_{\ucelc}$ is $\textsf{P}^{\textsf{NP}}_{\parallel}$-complete.	
	\begin{proof}
	In order to show $\textsf{P}^{\textsf{NP}}_{\parallel}$-completeness, we again use the characterisation of $\textsf{P}^{\textsf{NP}}_{\parallel}$ from \cite{Chang:1995} (see also Proposition~\ref{prop:exists:ucelc}). So we show that is $\textit{Cred}_{\ucelc}$ is $\textsf{P}^{\textsf{NP}}_{\parallel}$-complete by showing that
		\begin{enumerate}
			\item $\textit{Cred}_{\ucelc}\in \textsf{P}^{\textsf{NP}}_{\parallel}$,
			\item $\textit{Cred}_{\ucelc}$ is $\textsf{NP}$-hard,
			\item $\textit{Cred}_{\ucelc}$ is $\textsf{coNP}$-hard,
			\item Two problem instances $(\AF_1,a_1)$ and $(\AF_2,a_2)$ of $\textit{Cred}_{\ucelc}$ can be polynomially reduced to a problem instance $(\AF_3,a_3)$ of $\textit{Cred}_{\ucelc}$ such that $(\AF_3,a_3)$ is a positive instance if and only if both $(\AF_1,a_1)$ and $(\AF_2,a_2)$ are positive instances.
			\item A set of problem instances $I=\{(\AF_1,a_1),\ldots,(\AF_k,a_k)\}$ of $\textit{Cred}_{\ucelc}$ can be polynomially reduced to a problem instance $(\AF,a)$ of $\textit{Cred}_{\ucelc}$ such that $(\AF,a)$ is a positive instance if and only if there is at least one positive instance in $I$.
		\end{enumerate}
		We now show that properties 1--5 above hold:
		\begin{enumerate}
			\item For showing $\textsf{P}^{\textsf{NP}}_{\parallel}$-membership, we use a similar algorithm as in the proof of Proposition~\ref{prop:exists:ucelc}. Let $\AFcomplete$ be the input argumentation framework and $x$ the input argument..
		\begin{enumerate}
			\item[1.] For each argument $a\in\arguments$, check whether $a$ is not attacked by an initial set
			\item[2.] For each argument $a\in\arguments$, check whether $a$ is contained in an initial set
			\item[3.] Let $M$ be the set of arguments for which both checks 1 and 2 were positive
			\item[4.] Remove all unattacked arguments from $M$, yielding a new set $M'$
			\item[5.] Compute the maximal admissible set $M''$ in $M'$ (which is uniquely determined)
			\item[6.] If there is an initial set $S\subseteq M''$ with $x\in S$, return \textsc{Yes}, otherwise return \textsc{No}
		\end{enumerate}
		First observe that the above algorithm runs in $\textsf{P}^{\textsf{NP}}_{\parallel}$ using two consecutive rounds of parallel calls (which is still in $\textsf{P}^{\textsf{NP}}_{\parallel}$ due to Proposition~2.1 of \cite{Eiter:1997}): steps 1--5 run in $\textsf{P}^{\textsf{NP}}_{\parallel}\subseteq \textsf{P}^{\textsf{NP}}$ as already shown in the proof of Proposition~\ref{prop:exists:ucelc}, and step 6 can be solved by one further \textsf{NP}-oracle call (non-deterministically guess $S\subseteq M''$ and verify that it is an initial set with $x\in S$). The above algorithm also solves $\textit{Cred}_{\ucelc}$ as $M''$ contains all (and only) unchallenged initial sets.
		\item We show \textsf{DP}-hardness (which entails \textsf{NP}-hardness) instead. For that, we use the same reduction from $\textit{Unique}_{\st}$ as in the proof of Proposition~\ref{prop:unique:elc}, i.\,e., the construction $Tr_4$ from \cite{Dvorak:2011}, but first augment the input argumentation framework $\AFcomplete$ with a fresh argument $a$ (without any additional attacks), yielding an argumentation framework $\hat{\AF}=(\arguments\cup\{a\},\attacks)$. Note that $E$ is a stable extension of $\AF$ if and only if $E\cup\{a\}$ is a stable extension of $\hat{\AF}$. By the same argumentation as in the proof of Proposition~\ref{prop:unique:elc}, $\AF$ has a unique stable extension $E$ if and only if $Tr_4(\hat{\AF})$ has the unique (and unchallenged) initial set $E\cup\{a\}$. So $\AF$ has a unique stable extension if and only if $a$ is credulously accepted wrt.\ unchallenged initial sets in $Tr_4(\hat{\AF})$.	
		\item Since $\textit{Cred}_{\ucelc}$ is \textsf{DP}-hard (see above) it is also \textsf{coNP}-hard.
		\item Let $\AF_1=(\arguments_1,\attacks_1)$ and $AF_2=(\arguments_2,\attacks_2)$ be two argumentation frameworks and assume $\arguments_1\cap \arguments_2=\emptyset$ (otherwise rename arguments accordingly) and let $(AF_1,a_1)$ and $(AF_2,a_2)$ be two instances for $\textit{Cred}_{\ucelc}$. Construct an instance $(AF_3,a_3)$ with $\AF_3=(\arguments_3,\attacks_3)$ for $\textit{Cred}_{\ucelc}$ as follows (let $a_3$ be a fresh argument):
			\begin{align*}
				\arguments_3 & = \arguments_1\setminus\{a_1\}\cup \arguments_2\setminus\{a_2\} \cup\{a_3\}\\
				\attacks_3 & = \{(b,c)\in \attacks_1\mid b\neq a_1,c \neq a_1 \}\\
							& \quad \{(b,c)\in \attacks_2\mid b\neq a_2,c \neq a_2 \}\\
							& \quad \{(a_3,c) \mid (a_1,c)\in \attacks_1 \}\cup \{(b,a_3) \mid (b,a_1)\in \attacks_1 \}\\	
							& \quad \{(a_3,c) \mid (a_2,c)\in \attacks_2 \}\cup \{(b,a_3) \mid (b,a_2)\in \attacks_2 \}
			\end{align*}
			Informally speaking, $\AF_3$ is simply the union of $\AF_1$ and $\AF_2$ where the two arguments $a_1$ (from $\AF_1$) and $a_2$ (from $\AF_2$) are merged into a new argument $a_3$ that retains all previous attacks of $a_1$ and $a_2$. 
			
			We now show that $a_3$ is credulously accepted wrt.\ unchallenged initial sets in $\AF_3$ if and only if both $a_1$ and $a_2$ are credulously accepted wrt.\ unchallenged initial sets in $\AF_1$ and $\AF_2$, respectively. Without loss of generality, we assume that both $a_1$ and $a_2$ are attacked in $\AF_1$ and $\AF_2$, respectively (otherwise the problem trivialises since $\{a_1\}$ and/or $\{a_2\}$ are then unattacked initial sets). Assume now that $a_3$ is credulously accepted wrt.\ unchallenged initial sets in $\AF_3$ and let $M$ be an unchallenged initial set with $a_3\in M$. Consider $M_1= M\cap \arguments_1\cup \{a_1\}$ and observe
				\begin{itemize}
					\item $M_1$ is admissible in $\AF_1$: let $c\in \arguments_1$ be attacking $M_1$. Then $c$ also attacks $M$ in $\AF_3$ and $M$ defends itself in $\AF_3$ through some $d\in M$. Since there are no attacks between $\AF_1$ and $\AF_2$ either $d=a_3$ or $d\in \arguments_1$. In the first case, $a_1\in M_1$ then attacks $c$. In the second case $d\in M_1$ attacks $c$. It follows that $M_1$ is admissible in $\AF_1$.
					\item $M_1$ is an initial set in $\AF_1$: Assume there exists non-empty initial $M_1'\subsetneq M_1$. If $a_1\notin M_1'$ then $M_1'\subsetneq M$ as well, contradicting the fact that $M$ is initial. If $a_1\in M_1'$, it is also easy to see that $M_1'\setminus\{a_1\}\cup\{a_3\}\cup (M\cap \arguments_2)\subsetneq M$  must be an initial set of $\AF_3$.
					\item $M_1$ is unchallenged: Suppose there is another initial set $M_1'$ in conflict with $M_1$ in $\AF_1$. If $a_1\notin M_1'$ then $M_1'$ also is in conflict with $M$ in $\AF_3$, contradicting the fact that $M$ is unchallenged. The same follows for $a_1\in M_1'$ with the same argumentation as above.
				\end{itemize}
				For the same reason it follows that $M_2= M\cap \arguments_2\cup \{a_2\}$ is an unchallenged initial set in $\AF_2$ and, therefore, both $a_1$ and $a_2$ are credulously accepted wrt.\ unchallenged initial sets in $\AF_1$ and $\AF_2$, respectively. The other direction is analogous.
		\item Let $\AF_1=(\arguments_1,\attacks_1),\ldots, \AF_n=(\arguments_n,\attacks_n)$ be argumentation frameworks and assume $\arguments_i\cap \arguments_j=\emptyset$ for all $i,j=1,\ldots, n$ and $i\neq j$ (otherwise rename arguments accordingly). Let $I=\{(\AF_1,a_1),\ldots, (\AF_k,a_k)\}$ be a set of instances of $\textit{Cred}_{\ucelc}$. Construct $\AFcomplete$ as follows (let $a,b,c$ be fresh arguments):
			\begin{align*}
				\arguments & = \arguments_1\cup\ldots\cup\arguments_k\cup\{a,b,c\}	\\
				\attacks & = \attacks_1\cup\ldots\cup\attacks_k \cup\{(a,a),(a,b),(b,c),(c,c),(a_1,a),\ldots,(a_k,a),(c,a_1),\ldots, (c,a_k)\}
			\end{align*}
			A sketch of the construction is shown in Figure~\ref{fig:exreduct:or}.			
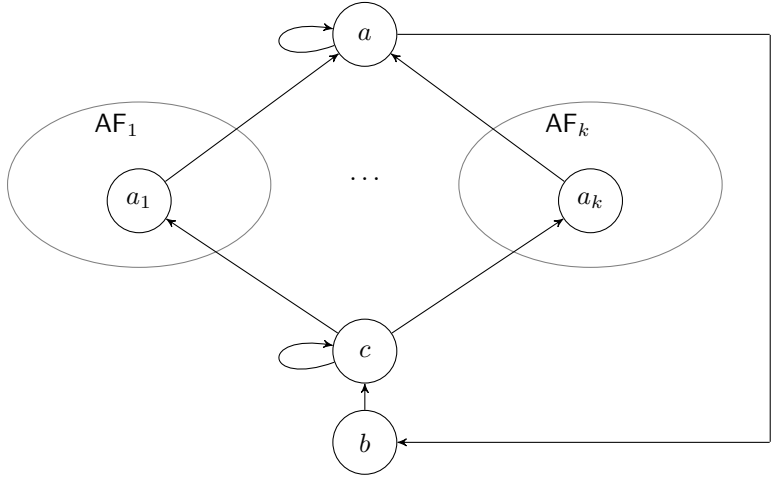
\begin{figure}[t]
\begin{center}
\begin{tikzpicture}[node distance=0.35cm]

	\node[args](a){$a$};
	
	\node[args, below=of a, yshift=-1cm, xshift=-3cm](a1){$a_1$};
	\node[below=of a, yshift=-1cm](ai){$\ldots$};
	\node[args, below=of a, yshift=-1cm, xshift=3cm](ak){$a_k$};
	
	\node[args, below=of a, yshift=-3cm](c){$c$};
		
	\node[ellipse,draw = gray,minimum width = 3.5cm, minimum height = 2.2cm] (af1) at (-3,-2) {};
	\node[ellipse,draw = gray,minimum width = 3.5cm, minimum height = 2.2cm] (afk) at (3,-2) {};
	\node(af1t) at (-3.3,-1.2) {$\AF_1$};
	\node(af1t) at (2.7,-1.2) {$\AF_k$};
	
	\node[args, below=of c](b){$b$};
	
	\path(a1) edge [->] (a);	
	\path(ak) edge [->] (a);	
	
	\path(c) edge [->] (a1);	
	\path(c) edge [->] (ak);	
	
	\path(b) edge [->] (c);	
	
	\path(a) edge [->,loop,in=170,out=200,min distance=10mm] (a);
	\path(c) edge [->,loop,in=170,out=200,min distance=10mm] (c);
		
	\node[right=of a, xshift=4.6cm, minimum size=0cm, inner sep=0cm](x1){};
	\node[right=of b, xshift=4.6cm, minimum size=0cm, inner sep=0cm](x2){};
	\path(a) edge [-] (x1);
	\path(x1) edge [-] (x2);
	\path(x2) edge [->] (b);

\end{tikzpicture}
\end{center}
\caption{A sketch of the argumentation framework $\AF$ from the proof of Proposition~\ref{prop:ucelc:cred}.}
\label{fig:exreduct:or}
\end{figure}	
				We now show that $b$ is credulously accepted wrt.\ unchallenged initial sets in $\AF$ if and only if there is $a_i$ ($i=1,\ldots,k$) that is credulously accepted wrt.\ unchallenged initial sets in $\AF_i$. Again, without loss of generality, we assume that all $a_i$  ($i=1,\ldots,k$) are attacked in $\AF_i$, respectively. Assume that $b$ is credulously accepted wrt.\ unchallenged initial sets in $\AF$ and let $M$ be an unchallenged initial set with $b\in M$. Since $M$ is admissible and $b$ is attacked by $a$, there must be $a_i\in M$ for some $i=1,\ldots,k$. Let $M_i=M\setminus\{b\}$. Then $M_i$ is necessarily an unchallenged initial set in $\AF_i$ (if, e.\,g., $M_i$ is challenged in $\AF_i$ then $M$ would also be challenged in $\AF$). It follows that $a_i$ is credulously accepted wrt.\ unchallenged initial sets in $\AF_i$. The other direction is analogous.\qedhere
		\end{enumerate}
	\end{proof}
\end{proposition}

\begin{proposition}\label{prop:ucelc:skept}
	$\textit{Skept}_{\ucelc}$ is $\textsf{P}^{\textsf{NP}}_{\parallel}$-complete.		
	\begin{proof}
	In order to show $\textsf{P}^{\textsf{NP}}_{\parallel}$-completeness, we use a characterisation of $\textsf{P}^{\textsf{NP}}_{\parallel}$ from \cite{Frei:2022}, in particular Corollary~8. This allows us to show $\textsf{P}^{\textsf{NP}}_{\parallel}$-completeness of $\textit{Skept}_{\ucelc}$ by showing that
		\begin{enumerate}
			\item $\textit{Skept}_{\ucelc}\in \textsf{P}^{\textsf{NP}}_{\parallel}$,
			\item $\textit{Skept}_{\ucelc}$ is $\textsf{coDP}$-hard,
			\item A set of problem instances $I=\{(\AF_1,a_1),\ldots,(\AF_k,a_k)\}$ of $\textit{Skept}_{\ucelc}$ can be polynomially reduced to a problem instance $(\AF,a)$ of $\textit{Skept}_{\ucelc}$ such that $(\AF,a)$ is a positive instance if and only if all instances in $I$ are positive.
		\end{enumerate}
		We now show that properties 1--3 above hold:
		\begin{enumerate}
			\item
	For $\textsf{P}^{\textsf{NP}}_{\parallel}$-membership, we use a similar algorithm as in the proof of Proposition~\ref{prop:exists:ucelc}. Let $\AFcomplete$ be the input argumentation framework and $x$ the input argument.
		\begin{enumerate}
			\item[1.] For each argument $a\in\arguments$, check whether $a$ is not attacked by an initial set
			\item[2.] For each argument $a\in\arguments$, check whether $a$ is contained in an initial set
			\item[3.] Let $M$ be the set of arguments for which both checks 1 and 2 were positive
			\item[4.] Remove all unattacked arguments from $M$, yielding a new set $M'$
			\item[5.] Compute the maximal admissible set $M''$ in $M'$ (which is uniquely determined)
			\item[6.] If $M''=\emptyset$ return \textsc{Yes}
			\item[7.] Let $M_{x}=M''\setminus \{x\}$
			\item[8.] Let $M'_{x}$ be the maximal admissible set in $M_{x}$ (which is uniquely determined)
			\item[9.] If $M'_{x}=\emptyset$ return \textsc{Yes}, otherwise return \textsc{No}
		\end{enumerate}
		First observe that the above algorithm runs in $\textsf{P}^{\textsf{NP}}_{\parallel}$. Steps 1--6 run in $\textsf{P}^{\textsf{NP}}_{\parallel}$ as already shown in the proof of Proposition~\ref{prop:exists:ucelc}. Furthermore, steps 7--9 run in (deterministic) polynomial time (in particular, step 8 runs in polynomial time by leveraging a similar algorithm as in the proof of Lemma~\ref{lem:subadm}). 
		
		We now claim that the above algorithm returns \textsc{Yes} if and only if $x$ is skeptically accepted wrt.\ unchallenged initial sets in $\AF$.
		\begin{itemize}
			\item ``$\Rightarrow$'': Assume the algorithm returns \textsc{Yes}. First, assume the algorithm terminates in step 6. $M''=\emptyset$ means that $\AF$ does not contain any unchallenged initial set (see the proof of Proposition~\ref{prop:exists:ucelc}). Then all arguments (including $x$) are trivially skeptically accepted. Now, assume the algorithm terminates in step 9. Since $M''$ is non-empty it contains unchallenged initial sets $S_{1},\ldots, S_{n}$ (which are all unchallenged initial sets of $\AF$, cf. the proof of Proposition~\ref{prop:exists:ucelc}). Since $M'_{x}=\emptyset$ it follows $S_{i}\nsubseteq M_{x}$ for all $i=1,\ldots,n$. It follows $x\in S_{i}$ for all $i=1,\ldots,n$ and $x$ is skeptically accepted.
			\item ``$\Leftarrow$'': Assume $x$ is skeptically accepted wrt.\ unchallenged initial sets. Consider the following case differentiation:
				\begin{itemize}
					\item There is no unchallenged initial set: in that case $M''=\emptyset$ in step 6 (see again the proof of Proposition~\ref{prop:exists:ucelc}) and the algorithm returns \textsc{Yes} in step 6.
					\item There are unchallenged initial sets $S_{1},\ldots, S_{n}$: since $x\in S_{i}$ for all $i=1,\ldots,n$ we have that $M'_{x}=\emptyset$ (all unchallenged initial sets are ``broken'' by removing $x$). Then the algorithm returns \textsc{Yes} in step 9.
				\end{itemize}
		\end{itemize}
		\item For showing \textsf{coDP}-hardness, we provide a reduction from the problem $\neg\textit{Unique}_{\st}$, i.\,e., the problem of deciding whether an argumentation framework $\AF$ does \emph{not} have a unique stable extension (which is naturally \textsf{coDP}-complete as its complement $\textit{Unique}_{\st}$ is \textsf{DP-complete}). We use  a similar approach as in the proof of Proposition~\ref{prop:ucelc:cred}, see also the proof of Proposition~\ref{prop:unique:elc}, again using the construction $Tr_4$ from \cite{Dvorak:2011}. For an input argumentation framework $\AFcomplete$ and a fresh argument $a$, we construct 
		\begin{align*}
			\tilde{\AF}=(\arguments\cup\{a\},\attacks\cup\{(b,a)\mid b\in \arguments\})
		\end{align*}
		In other words, we add an argument $a$ and attacks from each original argument to $a$. Note that $E$ is a stable extension of $\AF$ if and only if $E$ is a stable extension of $\tilde{\AF}$. We now claim that an input argumentation framework $\AF$ does not possess a unique stable extension if and only if $a$ is skeptically accepted wrt.\ unchallenged initial sets in $\tilde{\AF}'$. 
		\begin{itemize}
			\item Assume $\AF$ has no stable extension. Then $Tr_4(\tilde{\AF})$ has no initial set (see the proof of Proposition~\ref{prop:unique:elc}) and also no unchallenged initial set. Then every argument (including $a$) is trivially skeptically accepted wrt.\ unchallenged initial sets.
			\item Assume $\AF$ has a unique stable extension $E$. Then $Tr_4(\tilde{\AF})$ has the unique (and unchallenged) initial set $E$ that does not contain $a$. So $a$ is not skeptically accepted, as desired.
			\item Assume $\AF$ has more than one stable extension. Then $Tr_4(\tilde{\AF})$ has the same sets as (challenged) initial sets (as all these sets attack each other) and no unchallenged initial set. Then every argument (including $a$) is trivially skeptically accepted wrt.\ unchallenged initial sets.
		\end{itemize}
		For the other direction, assume $a$ is skeptically accepted wrt.\ unchallenged initial sets. As observed above, this can only happen if there are no unchallenged sets and this can only happen if $\AF$ does not have a unique stable extension. This shows \textsf{coDP}-hardness of $\textit{Skept}_{\ucelc}$.
		\item Let $\AF_1=(\arguments_1,\attacks_1),\ldots, \AF_n=(\arguments_n,\attacks_n)$ be argumentation frameworks and assume $\arguments_i\cap \arguments_j=\emptyset$ for all $i,j=1,\ldots, n$ and $i\neq j$ (otherwise rename arguments accordingly). Let $I=\{(\AF_1,a_1),\ldots, (\AF_k,a_k)\}$ be a set of instances of $\textit{Skept}_{\ucelc}$. We again assume that each $a_i$ is attacked in each $\AF_i$. Consider the argumentation framework $\AF$ following the construction in the proof of Proposition~\ref{prop:ucelc:cred}, item (5), see also Figure~\ref{fig:exreduct:or}. We claim that $b$ is skeptically accepted wrt.\ unchallenged initial sets in $\AF$ if and only if $a_1,\ldots,a_k$ are skeptically accepted wrt.\ unchallenged initial sets in $\AF_1,\ldots,\AF_k$, respectively. Without loss of generality, assume that $a_1$ is not skeptically accepted wrt.\ unchallenged initial sets in $\AF_1$. Then there is an unchallenged initial set $M$ in $\AF_1$ with $a_1\notin M$. Observe that $M$ is also necessarily an unchallenged initial set in $\AF$ (since $a_1\notin M$ there is no interaction with the rest of $\AF$). Since $b\notin M$ it follows that $b$ is also not skeptically accepted wrt.\ unchallenged initial sets in $\AF$. The argument generalises naturally to all $i=1,\ldots,k$. The other direction is analogous.\qedhere
		\end{enumerate}		
	\end{proof}
\end{proposition}

\begin{proposition}\label{prop:ver:celc}
	$\textit{Ver}_{\celc}$ is \textsf{NP}-complete.	
	\begin{proof}
		For \textsf{NP}-membership, on input $\AF$ and $S_{1}$ we guess a set $S_{2}$ that attacks $S_{1}$ and verify in polynomial time that $S_{2}$ is an initial set, cf.\ Proposition~\ref{prop:elc:poly}. This shows that $S_{1}$ is a challenged initial set.
		
		To show \textsf{NP}-hardness, we use the same reduction as in the proof of Proposition~\ref{prop:ver:ucelc} but using an instance of 3SAT. It is easy to see that an input instance $\phi$ is satisfiable if and only if $\{\psi\}$ is a challenged initial set in $\AF'_{\phi}$.
	\end{proof}
\end{proposition}	

\begin{proposition}
	$\textit{Exists}_{\celc}$ is \textsf{NP}-complete.	
	\begin{proof}
		For \textsf{NP}-membership, we guess two sets $S_{1}$ and $S_{2}$ that attack each other and verify in polynomial time that both are initial sets, cf.\ Proposition~\ref{prop:elc:poly}. This shows actually that both $S_{1}$ and $S_{2}$ are challenged initial sets.
		
		To show \textsf{NP}-hardness, we use the same reduction as in the proof of Proposition~\ref{prop:ver:ucelc} but using an instance of 3SAT. It is easy to see that an input instance $\phi$ is satisfiable if and only if there is a challenged initial set (concretely, $\{\psi\}$) in $\AF'_{\phi}$.
	\end{proof}
\end{proposition}

\begin{proposition}
	$\textit{Cred}_{\celc}$ is \textsf{NP}-complete.	
	\begin{proof}
		For \textsf{NP}-membership, on input $\AF$ and $a$ we guess two sets $S_{1}$ and $S_{2}$ with $a\in S_{1}$, $S_{1}$ and $S_{2}$ attack each other and verify in polynomial time that both are initial sets, cf.\ Proposition~\ref{prop:elc:poly}. This shows actually that both $S_{1}$ and $S_{2}$ are challenged initial sets and $a$ is credulously accepted wrt\ challanged initial sets (as $a\in S_{1}$).
		
		To show \textsf{NP}-hardness, we use the same reduction as in the proof of Proposition~\ref{prop:ver:ucelc} but using an instance of 3SAT. It is easy to see that an input instance $\phi$ is satisfiable if and only if $\psi$ is credulously accepted wrt.\ challenged initial sets in $\AF'_{\phi}$.
		\end{proof}
\end{proposition}

\begin{proposition}
	$\textit{Skept}_{\celc}$ is \textsf{coNP}-complete.		
	\begin{proof}
		For \textsf{coNP}-membership, we show that the complement problem, i.\,e., the problem of deciding whether an input argument $a$ is \emph{not} skeptically accepted wrt.\ initial sets, is in \textsf{NP}. For that we guess two sets $S_{1}$ and $S_{2}$ with $a\notin S_{1}$, $S_{1}$ and $S_{2}$ attack each other and verify in polynomial time that both are initial sets, cf.\ Proposition~\ref{prop:elc:poly}. This shows that $a$ is not skeptically accepted and therefore $\textit{Skept}_{\celc}$ is in \textsf{coNP}.
		
		To show \textsf{coNP}-hardness, we use the same reduction as in the proof of Proposition~\ref{prop:ver:ucelc}. It is easy to see that an input instance $\phi$ is unsatisfiable if and only if $\psi$ is skeptically accepted wrt.\ challenged initial sets in $\AF'_{\phi}$.
	\end{proof}
\end{proposition}


%

\end{document}